\documentclass[10pt,journal]{IEEEtran}

\usepackage{amsthm}
\usepackage{amsmath}
\usepackage{slashbox}
\usepackage{graphicx}
\usepackage{epsfig,graphics,subfigure,psfrag,amsmath,amssymb}
\usepackage{cite}
\usepackage{float}
\usepackage{tocvsec2}
\usepackage{color}
\usepackage{verbatim}
\usepackage{algorithm, algorithmic}
\usepackage{mathtools}
\usepackage{hhline}
\usepackage{enumerate}
\usepackage{arydshln}
\usepackage{bbm}
\usepackage{stackengine}
\usepackage{mathrsfs}
\usepackage{enumitem}
\usepackage{caption}
\usepackage{stfloats}
\usepackage{hyperref}

\usepackage{setspace}

\newtheorem{thm}{Theorem}
\newtheorem{lem}{Lemma}
\newtheorem{prop}{Proposition}

\newtheorem{rem}{Remark}

\newtheorem{assump}{Assumption}
\newtheorem{fact}{Fact}

\makeatletter

\newcommand{\Rmnum}[1]{\expandafter\@slowromancap\romannumeral #1@}
\makeatother

\makeatletter
\newcommand*\bigcdot{\mathpalette\bigcdot@{.5}}
\newcommand*\bigcdot@[2]{\mathbin{\vcenter{\hbox{\scalebox{#2}{$\m@th#1\bullet$}}}}}
\makeatother

\def\@#1{\pmb{#1}}
\def\bf#1{\mathbf{#1}}
\def\b#1{\mathbb{#1}}
\def\s#1{\mathsf{#1}}
\def\w#1{\widetilde{#1}}
\def\o#1{\overline{#1}}

\def\ca#1{\mathcal{#1}}

\newcommand\aleq{\mathrel{\stackrel{\makebox[0pt]{\mbox{\normalfont\tiny (a)}}}{\leq}}}
\newcommand\bleq{\mathrel{\stackrel{\makebox[0pt]{\mbox{\normalfont\tiny (b)}}}{\leq}}}

\newcommand\aeq{\mathrel{\stackrel{\makebox[0pt]{\mbox{\normalfont\tiny (a)}}}{=}}}

\DeclareMathOperator*{\argmin}{arg\,min}

\allowdisplaybreaks

\begin{document}
\title{{Decentralized Multi-Task Online Convex Optimization Under Random Link Failures}}
\author{Wenjing~Yan,~\IEEEmembership{Graduate~Student~Member,~IEEE}, and Xuanyu~Cao,~\IEEEmembership{Senior~Member,~IEEE} 
\thanks{This work was supported by the National Natural Science Foundation of China Grant 62203373. \emph{(Corresponding author: Xuanyu~Cao.)}

W. Yan and X. Cao are with the Department of Electronic and Computer Engineering, The Hong Kong University of Science and Technology, Hong Kong (e-mail: wj.yan@connect.ust.hk, eexcao@ust.hk). 
}}


\maketitle

\begin{abstract}
	Decentralized optimization methods often entail information exchange between neighbors. Transmission failures can happen due to network congestion, hardware/software issues, communication outage, and other factors. In this paper, we investigate the random link failure problem in decentralized multi-task online convex optimization, where agents have individual decisions that are coupled with each other via pairwise constraints. Although widely used in constrained optimization, conventional saddle-point algorithms are not directly applicable here because of random packet dropping. To address this issue, we develop a robust decentralized saddle-point algorithm against random link failures with heterogeneous probabilities by replacing the missing decisions of neighbors with their latest received values. Then, by judiciously bounding the accumulated deviation stemming from this replacement, we first establish that our algorithm achieves $\ca{O}(\sqrt{T})$ regret and $\ca{O}(T^\frac{3}{4})$ constraint violations for the full information scenario, where the complete information on the local cost function is revealed to each agent at the end of each time slot. These two bounds match, in order sense, the performance bounds of algorithms with perfect communications. Further, we extend our algorithm and analysis to the two-point bandit feedback scenario, where only the values of the local cost function at two random points are disclosed to each agent sequentially. Performance bounds of the same orders as the full information case are derived. Finally, we corroborate the efficacy of the proposed algorithms and the analytical results through numerical simulations. 
\end{abstract}


\section{Introduction}

Online convex optimization (OCO) has received increasing research attention in recent years \cite{hazan2016introduction,li2022survey,arjevani2020tight,cao2021decentralized1}, thanks to its broad applicability in various fields such as sensor networks, robotics, machine learning, smart grids, portfolios, and recommendation systems \cite{hazan2016introduction,li2022survey}. OCO involves minimizing a sequence of time-dependent cost functions, which are unknown beforehand and only revealed after the current decisions are made. The objective is to achieve sublinear regret, which measures the gap between the accumulated cost of the sequential functions incurred by the algorithm and the benchmark associated with the best-fixed decisions in hindsight. Initially, OCO was studied in a centralized fashion \cite{shalev2012online}, where the entire optimization task is conducted at a central entity with access to all agents' information. Recently, with the emergence of large-scale networks and the prevalence of big data in modern life, centralized optimization has been challenging to implement, motivating a growing body of research on optimization through direct collaboration among agents, which is the paradigm of decentralized optimization.

Decentralized optimization methods often entail information exchange between neighbors. Most existing studies \cite{yan2020passive,shalev2012online,cao2021decentralized1,yan2023zero,koppel2015saddle,yan2019passive} presume an ideal communication model that provides reliable and lossless transmission links. However, real-world communication systems may suffer from link failures caused by network congestion, hardware/software issues, communication outage (especially in harsh wireless environments) \cite{bonaventure2011computer}, and other factors. The conventional solution is to use transmission control protocol (TCP) \cite{postel1981transmission}, where the transmission failures are detected and recovered by acknowledgment (ACK), retransmission, and error detection mechanisms. The reliability provided by TCP comes with a price of prolonged latency since the convergence speed of decentralized algorithms is dominated by stragglers. Moreover, the communication overhead introduced by ACK and retransmission is often considerable. As a result, TCP may not be favorable for some time-stringent tasks that prioritize time over reliability, or when communication resources are too scarce to afford the TCP overhead. Instead, user datagram protocol (UDP) \cite{postel1980user} provides connectionless and unreliable packet delivery service by getting rid of these protection mechanisms, which can be preferable for certain cases. Nevertheless, UDP is at risk of link failures. The convergence behavior of decentralized optimization methods employing UDP needs to be understood.

In the literature, a bulk of studies have investigated the performance of decentralized algorithms under random link failures. Hatano and Mesbahi \cite{hatano2005agreement} first studied the decentralized consensus problem over random networks, and the asymptotic agreement throughout the network was demonstrated via notions of stochastic stability. Afterward, the gossip algorithm was proposed in \cite{boyd2006randomized} for the decentralized consensus in a randomly connected network. Results in \cite{boyd2006randomized} showed that the gossip algorithm's convergence time depends on the second largest eigenvalue of the doubly stochastic weight matrix. 
For the problem of consensus-based distributed optimization under random link failures, where the goal is to minimize the sum of the local cost functions with a common decision variable, Nedic and Ozdaglar \cite{nedic2009distributed} proposed an averaging-based sub-gradient descent algorithm, whose convergence relies on the double stochasticity of time-varying weight matrices. The methods of push-sum \cite{mai2016distributed} and subgradient-push \cite{xi2018linear} relax this requirement on weight matrices to column-stochastic. Furthermore, decentralized federated learning with unreliable communication was investigated in \cite{ye2022decentralized} under a doubly-stochastic weight matrix.

All the aforementioned studies on decentralized optimization under link failures are focused on consensus-based problems where the entire network aims to reach an agreement on decisions. However, in real-world networks, the decisions of neighboring agents can be related but different, raising the issue where consensus optimization techniques are not applicable. For instance, in multi-robot path planning, robots navigate toward their respective destinations while the positions of adjacent robots are subject to distance constraints \cite{bhattacharya2011distributed}. In distributed multi-task learning, different agents deal with similar tasks, and the trained models of neighboring agents have some correlations \cite{bhattarai2016cp}.

This paper studies decentralized multi-task OCO under random link failures, where agents have individual decisions that are coupled with each other via pairwise constraints. The formulation encompasses consensus optimization as a special case but allows for a more general hypothesis that there is data heterogeneity across the network \cite{koppel2017proximity,bedi2019asynchronous}. Pairwise constraints are commonly used in decentralized multi-robot systems to capture inter-robot relationships and facilitate cooperative behaviors among robots, such as maintaining a specific formation, avoiding collisions, and sharing information and resources \cite{bhattacharya2011distributed,ho2020decentralized}. Additionally, pairwise constraints are utilized for simultaneous localization and mapping \cite{mirowski2013signalslam}, distributed multi-robot trajectory \cite{salvado2022dimopt}, and the separation of arrival times in resource-constrained multi-robot missions \cite{fouad2020energy}.

Compared to consensus-based optimization, non-consensus optimization presents unique challenges when dealing with link failures. In consensus-based optimization, decisions from neighbors are typically handled by weighted averages, where the common solution for link failures is by allocating the weights of the missing decisions to the decision of the agent itself \cite{nedic2009distributed,mai2016distributed,xi2018linear}. Convergence of the resulting algorithm can be analyzed by referring to techniques developed for decentralized optimization under time-varying networks \cite{nedic2017achieving}. In contrast, in the considered decentralized multi-task OCO, neighbors' decisions are necessitated for computing the pairwise constraints. Since decisions vary with agents, any transmission failure of a neighbor makes the computation of the pairwise constraint impossible, which presents significant challenges for algorithm design and theoretical analysis.

To solve this problem, we develop a robust saddle-point algorithm for decentralized multi-task OCO against random link failures with heterogeneous failure probabilities across the network. Our algorithm allows for directional link failures, meaning that the failure of transmission in one direction of a communication edge does not imply the failure of its reverse direction. This is a more practical consideration than the unidirectional link failure assumption made in \cite{chen2021convergence} for consensus problems. In the development of our algorithm, we first introduce a substitute for the missing decisions by their latest received values. Based on these substitutes, we design approximated gradients for both the primal and dual variables. The approximation entails the pairwise constraints, the gradients of the pairwise constraints, and the dual variables. 

Performance analysis in our paper is more challenging than previous work \cite{cao2020decentralized} that studied quantized communication for the considered decentralized multi-task OCO model. In \cite{cao2020decentralized}, stochastic gradients are available, and the main variation of its algorithm from the perfect communication case \cite{koppel2017proximity} is the additional quantization error, which is independent of other quantities. However, in our paper, the approximated gradients are biased and cannot be eliminated by simply taking expectation. The biased approximated gradients cause deviations on the primal and dual variables through gradient update. Therefore, our theoretical analysis needs to quantify the deviations of these quantities throughout the entire algorithm iterations.

Another challenge in the analysis of our paper is that, due to the biased gradient approximation, the symmetry of the pairwise constraints is violated, and consequently, the symmetry of the dual variables. The same issue arises in another paper \cite{bedi2019asynchronous}, which investigated asynchronous communication for the same problem model under study. To address this issue, \cite{bedi2019asynchronous} requires the exchange of dual variables at each iteration. This incurs additional communication overhead. In contrast, our paper utilizes variable approximations and avoids the need to transmit these dual variables. Thus, our algorithm is more communication efficient than the approach of \cite{bedi2019asynchronous}.

Despite the aforementioned challenges, by judiciously quantifying the deviations of all quantities induced by gradient approximation, we establish that our algorithm achieves $\ca{O}(\sqrt{T})$ regret and $\ca{O}(T^\frac{3}{4})$ constraint violations for the full information scenario, where the complete information on the local cost function is revealed to each agent at the end of each time slot. These two bounds match, in order sense, the bounds of saddle-point algorithms with perfect communications \cite{koppel2017proximity}. 

Further, we extend our algorithm and performance analysis to the bandit feedback scenario, where only the values of the local cost function at two random points are disclosed to each agent at each time slot. Bandit feedback is prevalent in the real world, where full information on the cost functions is unavailable or difficult to obtain, and one only gets access to values of the cost functions at certain points, for instance, experiments in physics/chemistry/biology, learning from recommender system logs, and designs of Artificial Intelligence for games. By resorting to the gradient estimation based on pointwise values \cite{flaxman2005online}, we develop a modified saddle-point algorithm for the two-point bandit feedback scenario. Performance bounds of the same orders as the full information case are derived. Finally, the effectiveness of our algorithms and analytical results are validated through numerical experiments.

\emph{Notations:} We use $\b{R}$ to represent the set of real numbers and $\b{R}_+$ the set of non-negative real numbers. $[n]$ denotes the integer set $\{1,2,\cdots,n\}$. For a set $\ca{X}$, $\ca{X}^n$ represents the $n$-fold Cartesian product $\ca{X} \times\cdots\times\ca{X}$. The operator $|\cdot|$ over a set returns its cardinality. $\Pi_\ca{X}(\cdot)$ projects the input onto the set $\ca{X}$ and $[\cdot]^+$ projects the input onto the non-negative orthant. $\|\cdot\|_1$ and $\|\cdot\|$ are respectively the $\ell_1$ norm and $\ell_2$ norm. The operator $ \langle\cdot,\cdot \rangle $ takes the inner product on two input vectors of the same dimension. $\ca{N}(\mu,\sigma^2)$ stands for a Gaussian distribution with mean $\mu$ and variance $\sigma^2$.


\section{Full Information Scenario} \label{Sec_FI}

In this section, we study the decentralized multi-task OCO under random link failures with full information feedback, where the complete information on the local cost function is disclosed to each agent sequentially. We first develop a robust saddle-point algorithm against link failures with heterogeneous probabilities. Then, we analyze the convergence behavior of the proposed algorithm and establish sublinear regret and sublinear constraint violation bounds.

\subsection{Problem Formulation}

We consider an undirected graph $\ca{G}=(\ca{V}, \ca{E})$ with $|\ca{V}|=n$ agents and $ |\ca{E}|=\frac{m}{2}$ edges, where each edge corresponds to a communication link. We assume that bi-directional communication is allowed between each connected agent pair $(i, j)\in\ca{E}$. Denote by $\ca{N}_{i}$ the neighbor set of the $i$th agent. Suppose that the time is slotted. In each time slot $t$, agent $i$ decides its current action $\bf{x}_i^{(t)} \in\ca{X}$ with its current local cost function $f_{i}^{(t)}:\ca{X}\mapsto\b{R}$ unknown \emph{a priori}, where $\ca{X}\in\b{R}^d$ is the set of feasible decisions. The feasible sets across all agents are assumed to be the same. After $\bf{x}_i^{(t)}$ is chosen, $f_{i}^{(t)}$ is revealed to agent $i$ at the end of time slot $t$, i.e., \emph{full information} of the local cost function is accessible to the agent after a decision is made. The decisions of each connected agent pair are subject to a constraint $g_{ij}:\ca{X} \times \ca{X} \rightarrow \b{R} $ that $ g_{i j}\left(\bf{x}_{i}, \bf{x}_{j}\right) \leq 0 $, $\forall (i, j)\in\ca{E}$. We assume that the pairwise constraint functions are symmetric in parameters, i.e., $ g_{i j}\left(\bf{x}_{i}, \bf{x}_{j}\right)=g_{j i}\left(\bf{x}_{j},\bf{x}_{i}\right) $, $\forall (i, j)\in\ca{E}$. Further, the constraint functions are assumed to be fixed throughout the optimization procedure and available to all agents before the start of the optimization. Ideally, the goal of the decentralized multi-agent system is to minimize the total incurred cost over a time horizon $T$, i.e.,
\begin{align}
	\begin{split} 
		\min _{\bf{x}_i^{(t)} \in \ca{X},\forall i,t} ~&\sum_{t=1}^{T}\sum_{i=1}^{n} f_{i}^{(t)}\left(\bf{x}_{i}^{(t)}\right)  \\
		\text{subject to}  ~& \sum_{t=1}^{T}g_{i j}\left(\bf{x}_{i}^{(t)}, \bf{x}_{j}^{(t)}\right) \leq 0,
		\forall (i, j)\in\ca{E}. \label{Pro_def}
    \end{split}
\end{align}

In the literature, the most commonly used pairwise constraint is the quadratic constraint, which can model proximity and similarity/dissimilarity relationships among agents \cite{koppel2017proximity,bedi2019asynchronous,bhattarai2016cp}, and represents distance constraints in path planning \cite{bhattacharya2011distributed} and trajectory tasks \cite{bedi2019asynchronous}. Additionally, pairwise constraints can represent signal-to-interference-plus-noise restrictions in wireless communications, such as $\mathbb{E}\left[\operatorname{SINR}\left(\mathbf{x}_i, \mathbf{x}_j, \mathbf{h}_i\right)\right] \geq \gamma_{i j}$, where $\mathbf{x}_i$ is the signal send to device $i$ and $\mathbf{h}_i$ is the channle between device $i$ and the base station \cite{bedi2019asyn}. A budget constraint, $\gamma_{i j}^{\min } \leq x_i+x_j \leq \gamma_{i j}^{\max }$, can also be used in wireless communication for resource management (e.g., time or frequency) between each pair of transceivers \cite{sun2014joint}. In semi-supervised fuzzy clustering, the relative entropy constraint $D\left(\mathbf{x}_i \| \mathbf{x}_j\right) \leq \gamma_{i j}$ is used to exploit the intra-sample correlations between samples $\mathbf{x}_i$ and $\mathbf{x}_j$ for enhancing the clustering performance \cite{wang2023pairwise}. Moreover, the logistic constraint $\log \frac{\left|x_i-x_j\right|}{c} \geq \gamma_{ij}$ is also widely used, e.g., in \cite{fouad2020energy} to separate the arrival times of two successive robots for charging at the same station, where $x_i$ is the arrival time of robot $i$.

The formulation of Problem \eqref{Pro_def} encompasses a broad range of applications, such as multi-robot path planning \cite{bhattacharya2011distributed}, multi-task learning \cite{bhattarai2016cp}, correlated random field estimation \cite{koppel2017proximity,bedi2019asynchronous}, target localization with moving cameras \cite{bedi2019asynchronous}, and resource allocation in wireless communication systems \cite{bedi2019asyn}. Two examples illustrating multi-robot path planning and correlated random field estimation are elaborated in Section A of the supplementary file.

Problem \eqref{Pro_def} is, however, impossible to be solved in an online manner because it requires the non-causal information on future cost functions. A widely-adopted performance metric is static regret \cite{cao2020event,cao2021decentralized1}, given by
\begin{align*}
	\text{Reg}(T)=\sum_{t=1}^T\sum_{i=1}^nf_{i}^{(t)}(\bf{x}_i^{(t)})-\sum_{t=1}^T\sum_{i=1}^nf_{i}^{(t)}(\bf{x}_i^*),
\end{align*} 
where $\bf{x}^*=\left[\left(\bf{x}_1^*\right)^\s{T}, \cdots, \left(\bf{x}_n^*\right)^\s{T}\right]^\s{T}$  is the best decisions in hindsight, defined as 
\begin{align}\label{best_def}
	\begin{split}
	(\bf{x}_1^*,\cdots,\bf{x}_n^*)\in\underset{\bf{x}_i\in\ca{X}, \forall i}{\argmin}~&\sum_{t=1}^T\sum_{i=1}^nf_{i}^{(t)}(\bf{x}_i)\\
	\text{subject to}~& g_{i j}\left(\bf{x}_{i}, \bf{x}_{j}\right) \leq 0, 
	\forall (i, j)\in\ca{E}.
	\end{split}
\end{align}
Note that $\bf{x}^*$ cannot be computed online either and is only serving as a performance benchmark. Another performance metric of Problem \eqref{Pro_def} is constraint violation, given by
\begin{align*}
	\text{Vio}_{ij}(T)=\sum_{t=1}^T g_{ij}\left(\bf{x}_{i}^{(t)}, \bf{x}_{j}^{(t)}\right),\quad \forall (i, j)\in\ca{E}.
\end{align*}

Our goal is to design an algorithm that is robust against random link failures while achieving ``good'' performance. In general, an online algorithm is regarded “good” if both the regret and the constraint violations are sublinear with respect to (w.r.t.) $T$, i.e., $\text{Reg}(T)\leq o(T)$ and $\text{Vio}_{ij}(T)\leq o(T)$, for all $(i, j)\in\ca{E}$. Then, the time-average regret satisfies $\text{Reg}(T)/T\leq o(1)$, and the time-average constraint violations satisfy $\text{Vio}_{ij}(T)/T\leq o(1)$, for all $(i, j)\in\ca{E}$. Both are asymptotically non-positive as $T$ goes to infinity. Therefore, the asymptotic performance of the chosen actions $\left\{\bf{x}_i^{(t)}\right\}$ is no worse than that of the benchmark $\bf{x}^*$ in a time-average sense.

\subsection{Algorithm Development}

In this subsection, we develop a robust saddle-point algorithm for the decentralized multi-task OCO under random link failures. Our algorithm is based on finding a saddle point for the regularized Lagrangian of Problem \eqref{Pro_def}. The regularized Lagrangian of Problem \eqref{Pro_def} at time slot $t$ is defined as
\begin{align}
	\ca{L}^{(t)}(\bf{x}, \@{\lambda}) & = \sum_{i= 1}^{n}\left[f_{i}^{(t)}\left(\bf{x}_{i}\right) +\! \sum_{j \in \ca{N}_{i}}\!\left(\lambda_{i j} g_{i j}\left(\bf{x}_{i}, \bf{x}_{j}\right)\!-\!\frac{\delta \eta}{2} \lambda_{i j}^{2}\right)\!\right] \notag \\
	&= f^{(t)}(\bf{x})+ \left\langle\@{\lambda}, \bf{g}(\bf{x})\right\rangle -\frac{\delta \eta}{2}\|\@{\lambda}\|^{2}, \forall t \in[T],  \label{Equ_Lag}
\end{align}
where $\bf{x}=[\bf{x}_1^\s{T}, \cdots, \bf{x}_n^\s{T}]^\s{T}$ is the primal variable (agent decision); $f^{(t)}(\bf{x}) := \sum_{i= 1}^{n}f_{i}^{(t)}\left(\bf{x}_{i}\right)$ is the global cost function; $\lambda_{i j}$ is the dual variable (multiplier) associated with the constraint $g_{ij}(\cdot)\leq 0$, $\forall (i, j)\in\ca{E}$; $\@\lambda_i \in \b{R}_+^{|\ca{N}_i|}$ stacks the dual variables $\lambda_{ij}$, for all $j\in \ca{N}_i$; $\@\lambda = [\@\lambda_1^\s{T}, \cdots, \@\lambda_n^\s{T}]^\s{T}$; and $\bf{g}(\bf{x})$ is the concatenation of all the constraints. Note that in \eqref{Equ_Lag}, we add a regularizer $-\frac{\delta\eta}{2}\|\@\lambda\|^2$ to suppress the growth of the multiplier $\@\lambda$, so as to improve the stability of the algorithm. In \eqref{Equ_Lag}, $\eta>0$ is the stepsize of the algorithm, and $\delta>0$ is a control parameter.

To find the saddle-point of the Lagrangian in \eqref{Equ_Lag}, we utilize alternating gradient updates on the primal and dual variables.
The gradients of the Lagrangian $\ca{L}^{(t)}(\bf{x}, \@{\lambda})$ w.r.t. the primal variable $\bf{x}_i$ and the dual variable $\lambda_{i j}$ are respectively given by
\begin{align}
	\nabla_{\bf{x}_{i}} \ca{L}^{(t)}(\bf{x}, \@{\lambda}) =& \nabla_{\bf{x}_{i}} f_{i}^{(t)}\left(\bf{x}_{i}\right) +  \sum_{j \in \ca{N}_{i}}\left[\lambda_{i j} \nabla_{\bf{x}_{i}} g_{i j}\left(\bf{x}_{i}, \bf{x}_{j}\right) \right. \notag\\
	&\left. +\lambda_{j i} \nabla_{\bf{x}_{i}} g_{j i}\left(\bf{x}_{j}, \bf{x}_{i}\right)\right], \forall i\in[n], \label{primal_grad} \\
	\frac{\partial}{\partial \lambda_{i j}}\ca{L}^{(t)}(\bf{x}, \@{\lambda}) =& g_{i j}\left(\bf{x}_{i}, \bf{x}_{j}\right)-\delta \eta \lambda_{i j}, \forall (i, j)\in\ca{E}.  \label{dual_grad}
\end{align}
The computation of $\nabla_{\bf{x}_{i}} \ca{L}^{(t)}(\bf{x}, \@{\lambda})$ and $\frac{\partial}{\partial \lambda_{i j}}\ca{L}^{(t)}(\bf{x}, \@{\lambda})$ at agent $i$, $\forall i\in[n]$ requires the information on its neighbors' decisions $\bf{x}_{j}$, for all $j\in \ca{N}_i$. To this end, at each iteration of the decentralized saddle-point algorithm, all agents share their current decisions with their neighbors and simultaneously receive their neighbors' current decisions. However, transmission failures can occur during this process due to network congestion, hardware/software issues, communication outage, \cite{bonaventure2011computer} etc. The UDP solution addresses this problem via protection mechanisms, including ACK, retransmission, and error detection. This introduces additional communication overhead and causes prolonged latency. To improve communication efficiency, we adopt UDP in message transmission, which provides connectionless packet delivery but has no reliability guarantees. Thus, link failures need to be handled in the algorithm design.

To solve the link failure problem, we replace the missing decisions of neighbors with their latest received values. Specifically, at each agent $i\in[n]$, we store an auxiliary variable $\bf{x}_{j\to i}$ for each neighbor $j\in \ca{N}_i$. At each communication round $t>1$, if the transmission from agent $j$ to agent $i$ is successful, $\bf{x}_{j\to i}$ is updated by the currently received $\bf{x}_{j}^{(t)}$. Otherwise, $\bf{x}_{j\to i}$ remains unchanged. The failure probability of a transmission link usually depends on the channel quality, transceiver designs, and communication load of the network, which do not change significantly over the running time of an algorithm. Thus, we assume that the failure probabilities of all links are fixed throughout the time horizon $t\in[T]$. Denote by $p_{ij} \in[0,1)$ the failure probability of the transmission from agent $j$ to agent $i$, $\forall i\in[n], j\in\ca{N}_i$. For $t \geq 2$, we have
\begin{align} \label{Eq_prob}
	\bf{x}_{j\to i}^{(t)} = \begin{cases}
		\bf{x}_{j}^{(t)} & \text{with probability~} 1-p_{ij}, \\
		\bf{x}_{j\to i}^{(t-1)} &  \text{with probability~} p_{ij}.
	\end{cases}
\end{align}
Further, define $\o{p} := \max_{i,j}\{p_{ij}\}$ the maximum link failure probability. With these auxiliary variables $\left\{\bf{x}_{j\to i}\right\}$, we approximate the primal gradient in \eqref{primal_grad} as
\begin{align} \label{Approx_primal_grad}
	\bf{q}_i^{(t)} :=& \nabla_{\bf{x}_{i}} f_{i}^{(t)}\left(\bf{x}_{i}^{(t)}\right) \notag\\
	&+ 2\sum_{j \in \ca{N}_{i}} \lambda_{i j}^{(t)} \nabla_{\bf{x}_{i}} g_{i j}\left(\bf{x}_{i}^{(t)}, \bf{x}_{j\to i}^{(t)}\right),\forall i\in[n].
\end{align}
Similarly, we approximate the dual gradient in \eqref{dual_grad} as
\begin{align}
    r_{ij}^{(t)} :=& g_{i j}\left(\bf{x}_{i}^{(t)}, \bf{x}_{j\to i}^{(t)}\right) - \delta\eta\lambda_{ij}^{(t)},\forall (i, j)\in\ca{E}. \label{Approx_dual_grad}
\end{align}
Define $\bf{q}^{(t)} := \left[\left(\bf{q}_1^{(t)}\right)^\s{T}, \cdots, \left(\bf{q}_n^{(t)}\right)^\s{T}\right]^\s{T}$ and denote by $\bf{r}^{(t)}$ the concatenation of $\left\{r_{ij}^{(t)}\right\}$.

\begin{algorithm}[t]
    \caption{Robust Saddle-Point Algorithm for Decentralized Multi-Task OCO with Full Information: The Procedures at agent $i$, $\forall i\in[n]$.}
    \label{alg_FI}
    \begin{algorithmic}[1]
	\STATE Initialize $\bf{x}_{i}^{(1)} \in \ca{X}$ arbitrarily. Set $\lambda_{ij}^{(1)}=0$, $\forall j\in\ca{N}_i$. Receive $\bf{x}_{j}^{(1)}$ from neighbor $j$ and set $\bf{x}_{j\to i}^{(1)} = \bf{x}_{j}^{(1)}$, $\forall j\in\ca{N}_i$.
	\FOR{$t=2$ to $T$} 
	\STATE Send  $\bf{x}_{i}^{(t)}$ to neighbor $j$, $\forall j\in\ca{N}_i$.
	\STATE Receive $\bf{x}_{j}^{(t)}$ from neighbor $j$, $\forall j\in\ca{N}_i$.
	\STATE Update $\bf{x}_{j\to i}^{(t)}$, $\forall j\in\ca{N}_i$, as
	\begin{align*} 
		\bf{x}_{j\to i}^{(t)} = \begin{cases}
			\bf{x}_{j}^{(t)} & \text{if transmission succeeds}, \\
			\bf{x}_{j\to i}^{(t-1)} &  \text{if transmission fails}.
		\end{cases}
	\end{align*}
	\STATE Compute $\bf{q}_i^{(t)}$ by \eqref{Approx_primal_grad}.  
    \STATE Compute $r_{ij}^{(t)}$ by \eqref{Approx_dual_grad}, $\forall j\in\ca{N}_i$.
	\STATE Update the primal variable by
	\begin{align*} 
		\bf{x}_{i}^{(t+1)} =& \Pi_{\ca{X}}\left(\bf{x}_{i}^{(t)}-\eta \bf{q}_i^{(t)} \right).
	\end{align*}
	\STATE Update the dual variable by
	\begin{align*}
		\lambda_{i j}^{(t+1)} & = \left[\lambda_{i j}^{(t)} + \eta r_{ij}^{(t)}\right]^{+}, \forall j\in \ca{N}_i.
	\end{align*}
	\ENDFOR 
    \end{algorithmic}
\end{algorithm}

Based on the approximated gradients, we develop a robust saddle-point algorithm for the decentralized multi-task OCO against random link failures, which is presented in Algorithm \ref{alg_FI}.
In Algorithm \ref{alg_FI}, the initial agent decisions are randomly chosen from the admissible set $\ca{X}$. All the dual variables are initialized to $0$. We assume that communication at the first round is ensured, so that $\bf{x}_{j\to i}^{(1)} = \bf{x}_{j}^{(1)}$, $\forall i\in[n], j\in\ca{N}_i$. This can be achieved by employing TCP at the first round of transmission. In later iterations, each agent $i\in[n]$ sends its current decision $\bf{x}_{i}^{(t)}$ to its neighbors and receives $\bf{x}_{j}^{(t)}$ from all its neighbors $j\in\ca{N}_i$. Then, for any $j\in\ca{N}_i$, agent $i$ updates the value of $\bf{x}_{j\to i}^{(t)}$ based on whether $\bf{x}_{j}^{(t)}$ is successfully received or not. The primal variable $\bf{x}_i^{(t)}$ is updated using the approximated gradient $\bf{q}_i^{(t)}$ in Step 8, and the dual variable $\lambda_{i j}^{(t)}$ is updated using the approximated gradient $r_{ij}^{(t)}$ in Step 9, $\forall j\in\ca{N}_i$.

\subsection{Main Results}

In this subsection, we present our theoretical result on the regret and constraint violations of Algorithm \ref{alg_FI} in the presence of random link failures. First, we make the following assumptions.   
\begin{assump}[Compactness and convexity of $\ca{X}$] \label{assump_X}
	The set of admissible agent decisions $\ca{X}$ is closed, convex, and bounded, i.e., there exists a constant $R > 0$ such that $\|\bf{x}\| \leq R$, $\forall \bf{x} \in \ca{X}$.
\end{assump}
\begin{assump}[Convexity]\label{assump_convex}
	The local cost function $f_i^{(t)}(\bf{x}_i)$ is convex in $\bf{x}_i$, for any $\bf{x}_i \in \ca{X}$, $i\in [n]$, and $t\in[T]$. The pairwise constraint function $g_{ij}(\bf{x}_i, \bf{x}_j)$ is jointly convex in $\bf{x}_i$ and $\bf{x}_j$, for any $(i, j)\in\ca{E}$ and $ \bf{x}_i, \bf{x}_j \in \ca{X}$.
\end{assump}
\begin{assump}[Lipschitz continuity]\label{assump_f}
	There exists a constant $G>0$ such that for any $\bf{x}_{i}, \bf{x}_{i}^{\prime}\in \ca{X}$, $i \in[n]$, and $t\in[T]$,
	\begin{align*}
		\left|f_{i}^{(t)}\left(\bf{x}_{i}\right) - f_{i}^{(t)}\left(\bf{x}_{i}^{\prime}\right)\right|\leq G\|\bf{x}_{i}-\bf{x}_{i}^{\prime}\|.
	\end{align*}
\end{assump}	

Assumptions \ref{assump_X}-\ref{assump_f} are standard in the convergence analysis of constrained optimization \cite{koppel2017proximity,bedi2019asynchronous,cao2020event,yuan2017adaptive}, even in the absence of link failures. In addition, we make the following assumptions on the constraint functions $g_{ij}(\cdot)$ for any $(i, j)\in\ca{E}$.
\begin{assump}\label{assump_g_grad}
	There exists a constant $L>0$ such that for any $\bf{x}_i, \bf{x}_{i}^{\prime},\bf{x}_{j}, \bf{x}_{j}^{\prime}\in\ca{X}$ and $(i, j)\in\ca{E}$,
	\begin{align*}
		&\left\|\nabla_{\bf{x}_{i}} g_{i j}\left(\bf{x}_{i}, \bf{x}_{j}\right) - \nabla_{\bf{x}_{i}} g_{i j}\left(\bf{x}_{i}^{\prime}, \bf{x}_{j}\right)\right\|\leq L\left\|\bf{x}_{i} - \bf{x}_{i}^{\prime} \right\|,\\
		&\left\|\nabla_{\bf{x}_{i}} g_{i j}\left(\bf{x}_{i}, \bf{x}_{j}\right) - \nabla_{\bf{x}_{i}} g_{i j}\left(\bf{x}_{i}, \bf{x}_{j}^{\prime}\right)\right\|\leq L\left\|\bf{x}_{j} - \bf{x}_{j}^{\prime} \right\|. 
	\end{align*}
\end{assump}
\begin{assump}\label{assump_g}
    There exists a point $(\bf{x}_{i}^o,\bf{x}_{j}^o)\in\ca{X}^2$ such that for any $(i, j)\in\ca{E}$, $|g_{i j}\left(\bf{x}_{i}^o, \bf{x}_{j}^o\right)|$ is bounded. Moreover, there exists a point $(\bf{x}_{i}^o,\bf{x}_{j}^o)\in\ca{X}^2$ such that for any $(i, j)\in\ca{E}$, $\|\nabla_{\bf{x}_{i}}g_{i j}\left(\bf{x}_{i}^o, \bf{x}_{j}^o\right)\|$ is bounded.
\end{assump}
Assumption \ref{assump_g_grad} pertains to the Lipschitz continuity of the gradients of the constraint functions, which is commonly used in the constrained optimization problems \cite{koppel2017proximity,bedi2019asynchronous,cao2020decentralized,singh2024decentralized}. Assumption \ref{assump_g} is satisfied for practical functions. Given the boundedness of the set $\ca{X}$ (Assumption 1), numerous functions fulfill Assumptions \ref{assump_g_grad} and \ref{assump_g}. For instance, the constraint $\left\|\mathbf{x}_i - \mathbf{x}_j\right\|^2 - b_{ij} \leq 0$, which is frequently used to model the distance restriction in multi-robot path planning \cite{bhattacharya2011distributed,chen2020monopair} and the similarity constraint in linear/logistic regression \cite{liu2015pairwise}.  
\begin{prop}
	Under Assumptions \ref{assump_X}, \ref{assump_g_grad} and \ref{assump_g}, the constraint function $g_{ij}(\cdot)$, for any $(i, j)\in\ca{E}$, satisfies:
	\begin{enumerate} [label=(\roman*) ]
		\item  There exists a constant $\w{G}$ such that for any $\bf{x}_i, \bf{x}_{i}^{\prime},\bf{x}_j \in\ca{X}$, $\left|g_{i j}\left(\bf{x}_{i}, \bf{x}_{j}\right) - g_{i j}\left(\bf{x}_i^{\prime}, \bf{x}_{j}\right) \right|\leq \w{G}\left\|\bf{x}_{i} - \bf{x}_{i}^{\prime}\right\|$;
		\item There exists a constant $C$ such that for any $\bf{x}_i, \bf{x}_j \in\ca{X}$, $|g_{ij}(\bf{x}_{i}, \bf{x}_{j})| \leq C$.
	\end{enumerate}  \label{prop}
\end{prop}

The proof of Proposition \ref{prop} is presented in Section B of the supplementary file. With the above assumptions and Proposition \ref{prop}, we now present our theoretical result in Theorem \ref{t1} below for the regret and constraint violations of Algorithm \ref{alg_FI}. The detailed proof of Theorem \ref{t1} is presented in Section \ref{Sec_FI_Analysis}. 

\begin{thm} \label{t1}
    Define $\omega := \left(2+4m+ \gamma\right)\w{G}^2 + 8R^2L^2$ and $\gamma:= \max_{i} \left\{\sum_{j\in\ca{N}_{i}} \frac{20|\ca{N}_i|\left(1+\frac{1}{\beta}\right)p_{ji}}{1-(1+\beta)p_{ji}} \right\}$, where $\beta \in\left(0, \frac{1}{\o{p}}-1\right)$ is a positive constant. Set $\eta=\frac{a}{\sqrt{T}}$ with positive constant $a$. Choose $\delta\in \left[\frac{1-\sqrt{1-8\eta^2\omega}}{4 \eta^{2}}, \frac{1+\sqrt{1- 8\eta^2\omega}}{4 \eta^{2}}\right]$. 
	Then, with Assumptions \ref{assump_X}-\ref{assump_g}, for $T\geq 8a^2\omega$, the regret of Problem \eqref{Pro_def} is upper bounded by:
	\begin{align*}
		&\sum_{t= 1}^{T} \left(\b{E}\left[f^{(t)}\left(\bf{x}^{(t)}\right)\right] - f^{(t)}(\bf{x}^*)\right)  \\
		&\leq \left(\frac{2R^2}{a^2} + mC^2 + nG^2\right)a\sqrt{T} \\
		&~~~+\frac{5aG^2}{2} \sum_{i=1}^n \sum_{j\in\ca{N}_{i}} \frac{\left(1+\frac{1}{\beta}\right)p_{ij}}{1-(1+\beta)p_{ij}} \sqrt{T} = \ca{O}\left(\sqrt{T}\right).
	\end{align*}
	Further, for any $(i, j)\in\ca{E}$, the constraint violation is upper bounded by:
	\begin{align*}
		&\b{E}\left[\sum_{t=1}^{T} g_{i j}\left(\bf{x}_{i}^{(t)}, \bf{x}_{j}^{(t)}\right)\right] \\
		&\leq  2\sqrt{nGR} \left(\frac{1}{a} + a\delta + 2a\w{G}^2\right)^{\frac{1}{2}} T^{\frac{3}{4}} \\
		&~~+ \left(1+ a^2\delta + 2a^2\w{G}^2\right)^{\frac{1}{2}} \left( \frac{4R^2}{a^2} + 2mC^2 + 2nG^2 \right.\\
		&~~\left.+ 5G^2\sum_{i=1}^n \sum_{j\in\ca{N}_{i}} \frac{\left(1+\frac{1}{\beta}\right)p_{ij}}{1-(1+\beta)p_{ij}} \right)^{\frac{1}{2}}\!\!\sqrt{T} = \ca{O}\left(T^{\frac{3}{4}}\right). 
	\end{align*}
\end{thm}
\begin{rem} \label{rem_robust}
	Theorem \ref{t1} demonstrates that under random link failures with heterogeneous failure probabilities, Algorithm \ref{alg_FI} achieves $\ca{O}(\sqrt{T})$ regret and $\ca{O}(T^\frac{3}{4})$ constraint violations. Both bounds are sublinear and match, in order sense, the bounds of the saddle-point algorithm with perfect communications \cite{koppel2017proximity}. This verifies the robustness of our algorithm. More importantly, Theorem \ref{t1} manifests that the random link failures will not degrade the order of performance of our algorithm. 
\end{rem}
\begin{rem}
	In Theorem \ref{t1}, if we take the effect of link failure probabilities into consideration, the regret bound becomes $\ca{O}\left(\sum_{i=1}^n \sum_{j\in\ca{N}_{i}}\frac{p_{ij}}{1-(1+\beta)p_{ij}} \sqrt{T}\right)$. The quantity $\sum_{i=1}^n \sum_{j\in\ca{N}_{i}} \frac{p_{ij}}{1-(1+\beta)p_{ij}}$ is monotonic with all the link failure probabilities $\{p_{ij}\}$ that characterize the randomness/uncertainty of transmission links. It is evident that higher link failure probabilities lead to a larger regret bound. In particular, if link failures rarely happen, i.e., $p_{ij} \ll 1$, we have $\frac{p_{ij}}{1-(1+\beta)p_{ij}}\approx p_{ij}$, $\forall (i, j)\in\ca{E}$. Then, the regret bound can be approximated as $\ca{O}\left(\sum_{i=1}^n\sum_{j\in\ca{N}_{i}}p_{ij} \sqrt{T}\right)$, which is linear with the sum of all the link failure probabilities. Optimization algorithms can be designed to minimize the quantity $\sum_{i=1}^n \sum_{j\in\ca{N}_{i}} \frac{p_{ij}}{1-(1+\beta)p_{ij}}$, so as to tighten the regret bound. 
	
	For the constraint violations in Theorem \ref{t1}, the term that determines the performance order $\ca{O}(T^\frac{3}{4})$ is irrelevant to link failures. Thus, link failures will not directly affect the order of constraint violation bound of our algorithm. 
\end{rem}
\begin{rem}
	Despite achieving the same orders of regret and constraint violation bounds as the algorithm running in the perfect communication case \cite{koppel2017proximity}, the regret of Algorithm \ref{alg_FI} contains an extra coefficient $\sum_{i=1}^n \sum_{j\in\ca{N}_{i}} \frac{p_{ij}}{1-(1+\beta)p_{ij}}$ which is monotonically increasing w.r.t. the link failure probabilities $\{p_{ij}\}$. 
	This coefficient loosens the regret bound. Moreover, if transmission links have a high probability of failures (e.g., in a harsh wireless environment or with too stringent communication resources), the negative effect of this coefficient on the regret bound cannot be neglected. This demonstrates a tradeoff between communication conditions and optimization performance.
\end{rem}


\subsection{Performance Analysis} \label{Sec_FI_Analysis}


This subsection provides proof for Theorem \ref{t1}. First, we introduce a fact about a property of the projection operator.
\begin{fact} \label{Fact_projection}
	Suppose that set $\ca{A} \subset \b{R}^{d}$ is closed and convex. Then, for any $\bf{y} \in \b{R}^{d}$ and $\bf{x} \in \ca{A}$, we have
	\begin{align*} 
		\left\|\bf{x}-\Pi_{\ca{A}}(\bf{y})\right\| \leq\|\bf{x}-\bf{y}\|,
	\end{align*}
	where $\Pi_{\ca{A}}(\bf{y})$ denotes the projection of $\bf{y}$ onto the set $\ca{A}$ \cite{bertsekas2009convex}. 
\end{fact} 
Next, we present the following lemma related to the regularized Lagrangian in \eqref{Equ_Lag}.
\begin{lem}  \label{Lem_L-L}
	Consider the update steps in Algorithm \ref{alg_FI}. Under Assumptions \ref{assump_X}-\ref{assump_g}, for any $\bf{x} \in \ca{X}^{n}$, $\@{\lambda} \in \b{R}_+^m$, and $t\in[T]$, the Lagrangian in \eqref{Equ_Lag} satisfies:
	\begin{align} \label{Eq_Lem_1}
		&\ca{L}^{(t)}\left(\bf{x}^{(t)},\@{\lambda}\right) - \ca{L}^{(t)}(\bf{x},\@\lambda^{(t)}) \notag\\
		&\leq \frac{1}{2 \eta}\left(\left\|\@{\lambda} - \@{\lambda}^{(t)}\right\|^{2} - \left\|\@{\lambda} - \@{\lambda}^{(t+1)}\right\|^{2}\right) + \frac{\eta}{2}\left\|\bf{r}^{(t)}\right\|^{2} \notag\\
		&~~~+ \frac{1}{2 \eta}\left(\left\|\bf{x} - \bf{x}^{(t)}\right\|^{2} - \left\|\bf{x} - \bf{x}^{(t+1)}\right\|^{2}\right) + \frac{\eta}{2}\left\|\bf{q}^{(t)}\right\|^{2} \notag \\
		&~~~+ \left\langle \@{\lambda} - \@{\lambda}^{(t)}, \nabla_{\@{\lambda}} \ca{L}^{(t)}\left(\bf{x}^{(t)},  \@{\lambda}^{(t)}\right) - \bf{r}^{(t)}\right\rangle \notag \\
		&~~~+ \left\langle \bf{x}^{(t)} - \bf{x}, \nabla_{\bf{x}} \ca{L}^{(t)}\left(\bf{x}^{(t)}, \@{\lambda}^{(t)}\right) - \bf{q}^{(t)}\right\rangle.
	\end{align}
\end{lem}
The proof of Lemma \ref{Lem_L-L} is presented in Section C of the supplementary file. Lemma \ref{Lem_L-L} establishes a relationship between the Lagrangian in \eqref{Equ_Lag} and the primal and dual variables in Algorithm \ref{alg_FI}. In particular, in Eq.~\eqref{Eq_Lem_1}, the last two terms are introduced by the gradient errors, where $\nabla_{\@{\lambda}} \ca{L}^{(t)}\left(\bf{x}^{(t)},  \@{\lambda}^{(t)}\right) - \bf{r}^{(t)}$ is the gap between the true and the approximated dual gradient, and $\nabla_{\bf{x}} \ca{L}^{(t)}\left(\bf{x}^{(t)}, \@{\lambda}^{(t)}\right) - \bf{q}^{(t)}$ is the gap between the true and the approximated primal gradient. Next, we bound these two terms in Lemma \ref{Lem_grad_gap}.
\begin{lem}  \label{Lem_grad_gap}
	For any $t\in[T]$, we have 
	\begin{align*}
		1)~&\left\langle \@{\lambda} - \@{\lambda}^{(t)}, \nabla_{\@{\lambda}} \ca{L}^{(t)}\left(\bf{x}^{(t)},  \@{\lambda}^{(t)}\right) - \bf{r}^{(t)}\right\rangle \\
		&\leq \eta\w{G}^2 \left( \left\|\@{\lambda}\right\|^2 + \left\|\@{\lambda}^{(t)}\right\|^2 \right) + \frac{1}{4\eta} \sum_{i=1}^n\sum_{j\in\ca{N}_i} \epsilon_{ij}^{(t)};\\
		2)~& \left\langle \bf{x}^{(t)} - \bf{x}, \nabla_{\bf{x}} \ca{L}^{(t)}\left(\bf{x}^{(t)}, \@{\lambda}^{(t)}\right) - \bf{q}^{(t)}\right\rangle \\
		&\leq 4R^2L^2\eta\left\|\@{\lambda}^{(t)}\right\|^{2} + \frac{1}{\eta} \sum_{i=1}^n \sum_{j \in \ca{N}_{i}} \epsilon_{ij}^{(t)};
	\end{align*}
	where $\epsilon_{ij}^{(t)}:= \left\|\bf{x}_{j}^{(t)} -\bf{x}_{j\to i}^{(t)} \right\|^2$ is the error related to the transmission failure from agent $j$ to agent $i$ at the $t$th iteration, for any $i\in[n],j\in\ca{N}_i$ and $t\in[T]$.
\end{lem}
The proof of Lemma \ref{Lem_grad_gap} is presented in Section D of the supplementary file. Further, we bound the squares of $l_2$ norms of the approximated gradients $\bf{q}^{(t)}$ and $\bf{r}^{(t)}$ for any $ t\in[T]$ in the following lemma. 
\begin{lem} \label{Prop_grad_bound} 
	For any $t\in[T]$, the approximated primal and dual gradients used in Algorithm \ref{alg_FI} satisfy:
	\begin{enumerate} [label=(\alph*)]
		\item $\left\|\bf{q}_i^{(t)} \right\|^{2} \leq 2G^2 + 8|\ca{N}_i|\w{G}^2\left\|\@{\lambda}_i^{(t)}\right\|^{2}$, $\forall i\in[n]$;
		\item $\left\|\bf{q}^{(t)} \right\|^{2} \leq 2nG^2 + 4m\w{G}^2 \left\|\@{\lambda}^{(t)}\right\|^{2}$;
		\item $\left\|\bf{r}^{(t)} \right\|^{2} \leq 2mC^2 +2 \delta^{2} \eta^{2} \left\|\@{\lambda}^{(t)}\right\|^{2}$.
	\end{enumerate}
\end{lem} 
The proof of Lemma \ref{Prop_grad_bound} is presented in Section E of the supplementary file. Leveraging the results in Lemmas \ref{Lem_L-L}-\ref{Prop_grad_bound}, we have the following Lemma \ref{Lem_Lag_sum}.
\begin{lem}  \label{Lem_Lag_sum}
	For any $\bf{x} \in \ca{X}^{n}$ and $\@{\lambda} \in \b{R}_+^m$, the sum of the Lagrangian in \eqref{Equ_Lag} over $T$ iterations satisfies:
	\begin{align} \label{Eq_Lag_sum}
		&\sum_{t= 1}^{T} \b{E}\left[\ca{L}^{(t)}\left(\bf{x}^{(t)}, \@{\lambda}\right) - \ca{L}^{(t)}\left(\bf{x}^*,\@\lambda^{(t)}\right)\right] \notag\\
		&\leq\left(\frac{1}{2 \eta} + \eta T\w{G}^2\right) \left\|\@{\lambda}\right\|^2  +  \frac{2R^2}{\eta} + \eta T\left(mC^2 + nG^2 \right)\notag\\
		&~~~+ \frac{\eta}{2}\left(2\delta^{2} \eta^{2} + (2+4m)\w{G}^2 + 8R^2L^2 \right)\sum_{t= 1}^{T} \b{E}\left\|\@{\lambda}^{(t)}\right\|^{2} 	 \notag\\
		&~~~+ \frac{5}{4\eta} \sum_{i=1}^n \sum_{j\in\ca{N}_{i}} \sum_{t= 1}^{T}\b{E}\left[\epsilon_{ij}^{(t)} \right], 
	\end{align}
\end{lem}

\begin{figure*}[ht]
	\begin{align}  \label{Eq_bound1}
		&\sum_{t= 1}^{T} \left(\b{E}\left[f^{(t)}\left(\bf{x}^{(t)}\right)\right] - f^{(t)}(\bf{x}^*)\right) + \sum_{t = 1}^{T} \b{E}\left\langle\@{\lambda}, \bf{g}\left(\bf{x}^{(t)}\right)\right\rangle-\frac{\delta \eta T}{2}\|\@{\lambda}\|^{2} - \sum_{t = 1}^{T}\b{E}\left\langle\@{\lambda}^{(t)}, \bf{g}\left(\bf{x}^{*}\right)\right\rangle + \frac{\delta \eta}{2}\sum_{t = 1}^{T}\b{E}\|\@{\lambda}^{(t)}\|^{2} \notag\\
		&\leq \left(\frac{1}{2\eta} +\eta T\w{G}^2 \right)\left\|\@{\lambda}\right\|^2 + \frac{2R^2}{\eta} + \frac{\eta T}{2}\left(2mC^2 + 2nG^2 + 5G^2\sum_{i=1}^n \sum_{j\in\ca{N}_{i}} \frac{\left(1+\frac{1}{\beta}\right)p_{ij}}{1-(1+\beta)p_{ij}} \right)  \notag\\
		&~~~+ \frac{\eta}{2} \left(2\delta^{2} \eta^{2} + \left(2+4m\right)\w{G}^2 + 8R^2L^2 \right)\sum_{t= 1}^{T} \b{E}\left\|\@{\lambda}^{(t)}\right\|^{2} + \eta\w{G}^2 \sum_{i=1}^n \sum_{j\in\ca{N}_{j}} \frac{10|\ca{N}_j|\left(1+\frac{1}{\beta}\right)p_{ij}}{1-(1+\beta)p_{ij}} \sum_{t= 1}^{T} \b{E}\left\|\@{\lambda}_j^{(t)}\right\|^{2}. 
	\end{align}
	\begin{align} \label{Eq_bound2}
		&\sum_{t= 1}^{T} \left(\b{E}\left[f^{(t)}\left(\bf{x}^{(t)}\right)\right] - f^{(t)}(\bf{x}^*)\right) + \sum_{t=1}^{T}\b{E}\left\langle\@{\lambda}, \bf{g}\left(\bf{x}^{(t)}\right)\right\rangle - \frac{1}{2} \left(\frac{1}{\eta} + \eta T\left(\delta + 2\w{G}^2\right)\right) \|\@{\lambda}\|^{2} \notag\\
		&\leq \frac{2R^2}{\eta} + \frac{\eta T}{2}\left(2mC^2 + 2nG^2 + 5G^2\sum_{i=1}^n \sum_{j\in\ca{N}_{i}} \frac{\left(1+\frac{1}{\beta}\right)p_{ij}}{1-(1+\beta)p_{ij}} \right)
		+ \frac{\eta}{2} \left(2\delta^{2} \eta^{2} - \delta + \omega\right)\sum_{t= 1}^{T} \b{E}\left\|\@{\lambda}^{(t)}\right\|^{2} .
	\end{align}
	\rule[-10pt]{18.07cm}{0.1em}
\end{figure*}

The proof of Lemma \ref{Lem_Lag_sum} is presented in Section F of the supplementary file. Further, we have the following lemma about the error $\epsilon_{ij}^{(t)}$, for any $i\in[n],j\in\ca{N}_i$, and $t\in[T]$.
\begin{lem}  \label{Prop_error}
	For any $(i, j)\in\ca{E}$, the sum of the expected error $\b{E}\left[\epsilon_{ij}^{(t)} \right]$ over $T$ iterations satisfies:
	\begin{align*}
		&\sum_{t=1}^T \b{E}\left[\epsilon_{ij}^{(t)}\right] \notag\\
		&\leq \frac{\left(1+\frac{1}{\beta}\right)p_{ij}\eta^2}{1-(1+\beta)p_{ij}}  \left(2TG^2 \!+\! 8|\ca{N}_j|\w{G}^2 \sum_{t= 1}^{T} \b{E}\left\|\@{\lambda}_j^{(t)}\right\|^{2}\right),
	\end{align*}
	where $\beta \in\left(0, \frac{1}{\o{p}}-1\right)$ is a positive constant. 
\end{lem} 
The proof of Lemma \ref{Prop_error} is presented in Section G of the supplementary file.

\textbf{Proof of Theorem \ref{t1}:} Plugging the bound in  Lemma \ref{Prop_error} into \eqref{Eq_Lag_sum}, we obtain \eqref{Eq_bound1}, in which we use $\sum_{i=1}^n \sum_{j\in\ca{N}_{i}}\b{E}\left[\epsilon_{ij}^{(t)}\right] = \sum_{j=1}^n \sum_{i\in\ca{N}_{j}}\b{E}\left[\epsilon_{ij}^{(t)}\right]$. 
Further, the last term of \eqref{Eq_bound1} satisfies
\begin{align*} 
	&\eta\w{G}^2\sum_{i=1}^n \sum_{j\in\ca{N}_{i}} \frac{10|\ca{N}_i|\left(1+\frac{1}{\beta}\right)p_{ji}}{1-(1+\beta)p_{ji}}\sum_{t=1}^T \b{E}\left\|\@{\lambda}_i^{(t)}\right\|^{2} \notag\\
	&\leq \frac{\eta}{2} \w{G}^2 \gamma \sum_{t=1}^T \b{E}\left\|\@{\lambda}^{(t)}\right\|^{2}\!,
\end{align*}
where  $\gamma:= \max_{i} \left\{\sum_{j\in\ca{N}_{i}} \frac{20|\ca{N}_i|\left(1+\frac{1}{\beta}\right)p_{ji}}{1-(1+\beta)p_{ji}} \right\}$. Plugging into the above inequality and rearranging the terms in \eqref{Eq_bound1}, we obtain \eqref{Eq_bound2}, where $\omega := \left(2+4m+ \gamma\right)\w{G}^2 + 8R^2L^2$. Note that in \eqref{Eq_bound2}, we omit the term $\sum_{t = 1}^{T}\left\langle\@{\lambda}^{(t)}, \bf{g}\left(\bf{x}^{*}\right)\right\rangle$ since $\@{\lambda}^{(t)}\geq \bf{0}$ and $\bf{g}\left(\bf{x}^{*}\right)\leq \bf{0}$, for any $ t\in[T]$.

Next, we deal with the last term in \eqref{Eq_bound2}. We want to get rid of this term by properly choosing the stepsize $\eta$ and the parameter $\delta$, so that the coefficient $\frac{\eta}{2} \left(2\delta^{2} \eta^{2} - \delta + \omega\right)\leq 0$. Since $2\delta^{2} \eta^{2} - \delta + \omega$ is quadratic in $\eta$ and $\eta>0$, the following range of $\delta$ meets the desired inequality: 
\begin{align*} 
	&\delta \in\left[\frac{1-\sqrt{1-8\eta^2\omega}}{4 \eta^{2}}, \frac{1+\sqrt{1- 8\eta^2\omega}}{4 \eta^{2}}\right].
\end{align*}
We set $\eta=\frac{a}{\sqrt{T}}$, where $a> 0$ is a constant. To guarantee that the value of $\delta$ within the above interval is a real number, we require $1-8\eta^2\omega\geq 0$, i.e., the time horizon $T\geq 8a^2\omega$. 

Since any $\@{\lambda} \in \b{R}_+^m$ satisfies Eq. \eqref{Eq_bound2}, we set $\@{\lambda}=\frac{\left[\b{E}\left[\sum_{t=1}^{T} \bf{g}\left(\bf{x}^{(t)}\right)\right]\right]^+}{\frac{1}{\eta} + \eta T\left(\delta + 2\w{G}^2\right)}$. With $\delta$ chosen within the stated range above, we obtain 
\begin{align}
	&\sum_{t= 1}^{T} \left(\b{E}\left[f^{(t)}\left(\bf{x}^{(t)}\right)\right] - f^{(t)}(\bf{x}^*)\right) \notag\\
	&~~~+ \sum_{i=1}^{n} \sum_{j \in \ca{N}_{i}} \frac{\left(\left[\b{E}\left[\sum_{t=1}^{T} g_{i j}\left(\bf{x}_{i}^{(t)}, \bf{x}_{j}^{(t)}\right)\right]\right]^{+}\right)^{2}}{2\left(\frac{1}{\eta} + \eta T\left(\delta + 2\w{G}^2\right)\right)} \notag\\
	&\leq \frac{2R^2}{\eta} + \eta T\left(mC^2 + nG^2 \right)\notag\\
	&~~~ + \frac{5\eta TG^2}{2}\sum_{i=1}^n \sum_{j\in\ca{N}_{i}}\frac{\left(1+\frac{1}{\beta}\right)p_{ij}}{1-(1+\beta)p_{ij}} . \label{Eq_bound3}
\end{align}
The second term on the left side of \eqref{Eq_bound3} is non-negative and can be omitted. Then, plugging into $\eta=\frac{a}{\sqrt{T}}$, the regret bound in Theorem \ref{t1} is derived. 

Further, for any $\bf{x}\in\ca{X}^n$ and $t\in[T]$, the cost function $f^{(t)}(\cdot)$ satisfies
\begin{align} \label{Eq_OptGap} 
	\left|f^{(t)}\left(\bf{x}\right) - f^{(t)}\left(\bf{x}^*\right)\right|\leq G\|\bf{x}-\bf{x}^*\|\leq2nGR,
\end{align}
where the first inequality is based on the $G$-Lipschitz continuity of $f^{(t)}(\cdot)$, and the second inequality is from the boundedness of the set $\ca{X}$. Applying \eqref{Eq_OptGap} in \eqref{Eq_bound3}, we obtain
\begin{align} \label{Eq_sum_g}
	&\sum_{i=1}^{n} \sum_{j \in \ca{N}_{i}} \frac{\left(\left[\b{E}\left[\sum_{t=1}^{T} g_{i j}\left(\bf{x}_{i}^{(t)}, \bf{x}_{j}^{(t)}\right)\right]\right]^{+}\right)^{2}}{2\left(\frac{1}{\eta} + \eta T\left(\delta + 2\w{G}^2\right)\right)}  \notag\\
	&\leq 2nGRT + \frac{2R^2}{\eta} + \eta T\left(mC^2 + nG^2\right) \notag\\
	&~~~+ \frac{5\eta TG^2}{2}\sum_{i=1}^n \sum_{j\in\ca{N}_{i}} \frac{\left(1+\frac{1}{\beta}\right)p_{ij}}{1-(1+\beta)p_{ij}}. 
\end{align}
Note that for each given $\left(\left[\b{E}\left[\sum_{t=1}^{T} g_{i j}\left(\bf{x}_{i}^{(t)}, \bf{x}_{j}^{(t)}\right)\right]\right]^{+}\right)^{2}$ and $(i, j)\in\ca{E}$, the bound in \eqref{Eq_sum_g} also holds. Further, taking the square root on the both sides of \eqref{Eq_sum_g} and using the inequality $\sqrt{a+b+c} \leq \sqrt{a} + \sqrt{b} + \sqrt{c}$, $\forall a,b,c\geq 0$, we obtain 
\begin{align*}
	&\left[\b{E}\left[\sum_{t=1}^{T} g_{i j}\left(\bf{x}_{i}^{(t)}, \bf{x}_{j}^{(t)}\right)\right]\right]^{+} \notag\\
	&\leq 2\sqrt{nGRT}\left(\frac{1}{\eta} + \eta T\left(\delta + 2\w{G}^2\right)\right)^{\frac{1}{2}} \\
	&~~+ \left(T + \eta^2 T^2\left(\delta + 2\w{G}^2\right)\right)^{\frac{1}{2}} \left(\frac{4R^2}{\eta^2 T}  + 2mC^2 \right. \\
	&~~\left.+ 2nG^2 + 5G^2\sum_{i=1}^n \sum_{j\in\ca{N}_{i}} \!\!\frac{\left(1+\frac{1}{\beta}\right)p_{ij}}{1-(1+\beta)p_{ij}} \right)^{\frac{1}{2}}\!,
\end{align*}
where $\eta=\frac{a}{\sqrt{T}}$. Then, with the fact that $\b{E}\left[\sum_{t=1}^{T} g_{i j}\left(\bf{x}_{i}^{(t)}, \bf{x}_{j}^{(t)}\right)\right] \leq \left[\b{E}\left[\sum_{t=1}^{T} g_{i j}\left(\bf{x}_{i}^{(t)}, \bf{x}_{j}^{(t)}\right)\right]\right]^{+} $, the constraint violation bound in Theorem \ref{t1} is derived.


\section{Bandit Feedback Scenario}
\label{Sec_BF}

In this section, we focus on the bandit feedback scenario where each agent only has access to the values of its cost function at some random points in each time slot. We first formulate the bandit OCO problem and present some preliminaries pertaining to bandit optimization. Then, we modify the algorithm developed for the full information scenario in Section \ref{Sec_FI} for the two-point bandit feedback case. Finally, performance bounds of the same orders as the full information scenario are established. 

\subsection{Problem Formulation and Preliminaries}

Denote by $\b{S}:=\left\{\bf{u} \in \b{R}^{d} \mid\|\bf{u}\|=1\right\}$ the $d$-dimensional unit sphere and by $\b{B}:=\left\{\bf{u} \in \b{R}^{d} \mid\|\bf{u}\| \leq 1\right\}$ the $d$-dimensional unit ball. In two-point bandit feedback scenario, instead of full information, only the values of $f_{i}^{(t)}$ at the two points $\left(\bf{x}_i^{(t)}\pm\zeta\bf{u}_i^{(t)}\right)$ are revealed to each agent $i$ at the end of time slot $t$, where $\bf{x}_i^{(t)}$ is agent $i$'s decision at the $t$th time slot, $\zeta>0$ is a small constant, and $\bf{u}_i^{(t)}$ is a random vector sampled uniformly over the unit sphere $\b{S}$. In the bandit feedback scenario, the exact gradient $\nabla f_{i}^{(t)}\left(\bf{x}_i^{(t)}\right)$ is unavailable, and we have to estimate the gradient based on bandit feedback. Next, we provide some preliminaries \cite{flaxman2005online} about gradient estimation based on pointwise values. 
\begin{fact}\label{fact_gradest}
	Consider a function $\phi: \b{R}^{d} \rightarrow \b{R}$. Define $\w{\phi}(\bf{x}):=\b{E}_{\bf{v} \sim \ca{U}(\b{B})}[\phi(\bf{x}+\zeta \bf{v}) \bf{v}]$, where $\zeta>0$, and $\ca{U}(\b{B})$ denotes the uniform distribution over the unit ball $\b{B}\subset \b{R}^{d}$. The following statements hold:
	\begin{enumerate} [label=(\roman*) ]
		\item If $\phi$ is convex, so is $\w{\phi}$;
		\item If $\phi$ is $L$-Lipschitz continuous, so is $\w{\phi}$;
		\item For any  $\bf{x} \in \b{R}^{d}$:
		\begin{align*}
			\nabla_{\bf{x}} \w{\phi}(\bf{x})=\frac{d}{\zeta} \b{E}_{\bf{u} \sim \ca{U}(\b{S})}[\phi(\bf{x}+\zeta \bf{u}) \bf{u}],
		\end{align*}
		where $\ca{U}(\b{S})$ denotes the uniform distribution over the unit sphere $\b{S}\in \b{R}^d$.
	\end{enumerate}
\end{fact}

To ensure that the decisions $\bf{x}_{i} \pm \zeta \bf{u}_{i}$ are within the feasible set $\ca{X}$, we make an additional assumption on the topology of $\ca{X}$ below. Let $\ca{B}\left(\bf{z}_{0}, r\right)$ be the Euclidean ball centered at $\bf{z}_{0}$ with radius $r$, i.e., $\ca{B}\left(\bf{z}_{0}, r\right)=\left\{\bf{x} \mid\left\|\bf{x}-\bf{z}_{0}\right\| \leq r\right\}$.
\begin{assump}\label{assump_int}
	The set $\ca{X}$ has a non-empty interior, i.e., there exist $\bf{z}_{0} \in \ca{X}$ and $r>0$ such that $\ca{B}\left(\bf{z}_{0}, r\right) \subset \ca{X}$. 
\end{assump}

By Assumption \ref{assump_int}, we have $\ca{B}\left((1-\alpha) \bf{x}+\alpha \bf{z}_{0}, \alpha r\right) \subset \ca{X}$, $\forall \alpha \in(0,1)$ and $\bf{x} \in \ca{X}$. Set $\zeta<r$ and choose $\alpha \in\left[\frac{\zeta}{r},1\right)$. Then, with $\bf{x}_{i}$ taken from the set $\w{\ca{X}}:=\{(1-\alpha)\bf{x}+\alpha\bf{z}_0|\bf{x}\in\ca{X}\}$, $\bf{x}_{i} \pm \zeta \bf{u}_{i}\in \ca{X}$ is guaranteed, $\forall i\in[n]$ and $\bf{u}_i \sim \ca{U}(\b{B})$. By Assumption \ref{assump_X}, we know that $\w{\ca{X}}$ is a closed convex subset of $\ca{X}$. Further, we define $\w{f}_{i}^{(t)}:\w{\ca{X}}\mapsto\b{R}$ as
\begin{align*}
	\w{f}_{i}^{(t)}(\bf{x}_i):= \b{E}_{\bf{v}_i\sim\ca{U}(\b{B})}[f_{i}^{(t)}(\bf{x}_i+\zeta\bf{v}_i)], \forall i\in[n], t\in[T].
\end{align*}
By Fact \ref{fact_gradest} and Assumptions \ref{assump_convex} and \ref{assump_f}, we know that $\w{f}_{i}^{(t)}(\cdot)$ is convex and $G$-Lipschitz continuous, for any $i\in[n]$, $t\in[T]$. 

\subsection{Algorithm Development}


In this subsection, we develop a modified saddle-point algorithm for the decentralized multi-task OCO under random link failures with two-point bandit feedback. Since the gradient of $f_i^{(t)}(\cdot)$ is unavailable in the bandit feedback scenario, we substitute it with the gradient of $\w{f}_i^{(t)}(\cdot)$ for any $ i\in[n],t\in[T]$, which can be computed based on the results of Fact \ref{fact_gradest}. First, we define a new regularized Lagrangian corresponding to $\sum_{i= 1}^{n}\w{f}_{i}^{(t)}(\cdot)$ as   
\begin{align} 
	\w{\ca{L}}^{(t)}(\bf{x}, \@{\lambda}) & = \sum_{i= 1}^{n}\!\left[\w{f}_{i}^{(t)}\left(\bf{x}_{i}\right) \!+\! \sum_{j \in \ca{N}_{i}}\!\left(\!\lambda_{i j} g_{i j}\left(\bf{x}_{i}, \bf{x}_{j}\right)-\frac{\delta \eta}{2} \lambda_{i j}^{2}\right)\!\right] \notag \\
	&= \w{f}^{(t)}(\bf{x}) + \@{\lambda}^\s{T} \bf{g}(\bf{x})-\frac{\delta \eta}{2}\|\@{\lambda}\|^{2}, \forall t\in[T], \label{Equ_band_Lag}
\end{align}
where $\w{f}^{(t)}(\bf{x}) := \sum_{i= 1}^{n}\w{f}_{i}^{(t)}\left(\bf{x}_{i}\right)$. The gradients of $\w{\ca{L}}^{(t)}(\bf{x}, \@{\lambda})$ w.r.t. the primal and dual variables are respectively given by 
\begin{align}
	\nabla_{\bf{x}_{i}} \w{\ca{L}}^{(t)}(\bf{x}, \@{\lambda}) =& \nabla_{\bf{x}_{i}} \w{f}_{i}^{(t)}\left(\bf{x}_{i}\right) \!+\! \sum_{j \in \ca{N}_{i}}\left[\lambda_{i j} \nabla_{\bf{x}_{i}} g_{i j}\left(\bf{x}_{i}, \bf{x}_{j}\right) \right. \notag\\
	& \left. +\lambda_{j i} \nabla_{\bf{x}_{i}} g_{j i}\left(\bf{x}_{j}, \bf{x}_{i}\right)\right],\forall i\in[n],  \label{Eq_BF_PG}\\
	\frac{\partial}{\partial \lambda_{i j}}\w{\ca{L}}^{(t)}(\bf{x}, \@{\lambda}) =& g_{i j}\left(\bf{x}_{i}, \bf{x}_{j}\right)-\delta \eta \lambda_{i j}, \forall (i, j)\in\ca{E}. \notag  
\end{align}
By Fact \ref{fact_gradest}, for any $i\in[n]$ and $t\in[T]$, we have
\begin{align}
	\nabla_{\bf{x}_i} \w{f}_{i}^{(t)}\left(\bf{x}_{i}\right) =&  \b{E}_{\bf{u}_{i} \sim \ca{U}(\b{S})} \left[\frac{d}{2 \zeta}\left[f_i^{(t)}\left(\bf{x}_{i}+\zeta \bf{u}_{i}\right) \right.\right. \notag\\
	&\left.\left.\left. - f_i^{(t)}\left(\bf{x}_{i}-\zeta \bf{u}_{i}\right)\right] \bf{u}_{i}\right|\bf{x}_{i} \right]. \label{band_primal_grad}
\end{align}
By \eqref{band_primal_grad}, the quantity $\frac{d}{2 \zeta}\left[f_i^{(t)}\left(\bf{x}_{i}+\zeta \bf{u}_{i}\right)-f_i^{(t)}\left(\bf{x}_{i}-\zeta \bf{u}_{i}\right)\right]\bf{u}_{i}$ is a conditionally unbiased estimation of $\nabla_{\bf{x}_i} \w{f}_{i}^{(t)}\left(\bf{x}_{i}\right) $ for any given $\bf{x}_{i}$.

Gradient computation in the bandit feedback scenario suffers from the same link failure problem as in the full information case. To resolve this problem, we use the auxiliary variables $\left\{\bf{x}_{j\to i}\right\}$ again for gradient approximations. Similar to \eqref{Approx_primal_grad}, the primal gradient in \eqref{Eq_BF_PG} is approximated as 
\begin{align}  \label{Band_Approx_dual_grad}
	\w{\bf{q}}_i^{(t)} :&=\frac{d}{2 \zeta}\left[f_{i}^{(t)}\left(\bf{x}_{i}^{(t)}+\zeta \bf{u}_{i}^{(t)}\right)-f_{i}^{(t)}\left(\bf{x}_{i}^{(t)}-\zeta \bf{u}_{i}^{(t)}\right)\right] \bf{u}_{i}^{(t)} \notag\\
	&+\! 2\!\!\sum_{j \in \ca{N}_{i}} \!\lambda_{i j}^{(t)} \nabla_{\bf{x}_{i}} g_{i j}\left(\bf{x}_{i}^{(t)}, \bf{x}_{j\to i}^{(t)}\right),\forall i\in[n],t\in[T].
\end{align}
The approximation for the dual gradient remains 
\begin{align}
    r_{ij}^{(t)} :=& g_{i j}\left(\bf{x}_{i}^{(t)}, \bf{x}_{j\to i}^{(t)}\right) - \delta\eta\lambda_{ij}^{(t)},\forall (i, j)\in\ca{E}. \label{Approx_dual_grad2}
\end{align}

The modified saddle-point algorithm for the two-point bandit feedback scenario under random link failures is presented in Algorithm \ref{alg_BF}. The development of Algorithm \ref{alg_BF} is similar to that of Algorithm \ref{alg_FI}. 

\begin{algorithm}[t]
    \caption{Robust Saddle-Point Algorithm for Decentralized Multi-Task OCO with Two-Point Bandit Feedback: The Procedures at agent $i$, $\forall i\in[n]$.}
    \label{alg_BF}
    \begin{algorithmic}[1]
		\STATE Initialize $\bf{x}_{i}^{(1)} \in \w{\ca{X}}$ arbitrarily. Set $\lambda_{ij}^{(1)}=0$, $\forall j\in\ca{N}_i$. Receive $\bf{x}_{j}^{(1)}$ from neighbor $j$ and set $\bf{x}_{j\to i}^{(1)} = \bf{x}_{j}^{(1)}$, $\forall j\in\ca{N}_i$. 
	\FOR{$t=2$ to $T$} 
	\STATE Send  $\bf{x}_{i}^{(t)}$ to neighbor $j$, $\forall j\in\ca{N}_i$. 
	\STATE Receive  $\bf{x}_{j}^{(t)}$ from neighbor $j$, $\forall j\in\ca{N}_i$.
	\STATE Update $\bf{x}_{j\to i}^{(t)}$, $\forall j\in\ca{N}_i$, as
	\begin{align*} 
		\bf{x}_{j\to i}^{(t)} = \begin{cases}
			\bf{x}_{j}^{(t)} & \text{transmission succeeds} \\
			\bf{x}_{j\to i}^{(t-1)} &  \text{transmission fails}.
		\end{cases}
	\end{align*}
	\STATE Compute $\w{\bf{q}}_i^{(t)}$ by \eqref{Band_Approx_dual_grad}.  
    \STATE Compute $r_{ij}^{(t)}$ by \eqref{Approx_dual_grad2}, $\forall j\in\ca{N}_i$.
	\STATE Update the primal variable by
	\begin{align*} 
		\bf{x}_{i}^{(t+1)} =& \Pi_{\w{\ca{X}}}\left(\bf{x}_{i}^{(t)}-\eta \w{\bf{q}}_i^{(t)} \right).
	\end{align*}
	\STATE Update the dual variable by
	\begin{align*} 
		\lambda_{i j}^{(t+1)} & = \left[\lambda_{i j}^{(t)} + \eta r_{ij}^{(t)}\right]^{+}, \forall j\in \ca{N}_i.
	\end{align*}
	\ENDFOR 
    \end{algorithmic}
\end{algorithm}

\subsection{Main Results}

We now present the regret and constraint violations of Algorithm \ref{alg_BF} for the two-point bandit feedback scenario.
\begin{thm} \label{t2}
	Define $\omega := \left(2+4m+ \gamma\right)\w{G}^2 + 8R^2L^2$ and $\gamma:= \max_{i} \left\{\sum_{j\in\ca{N}_{i}} \frac{20\left(1+\frac{1}{\beta}\right)p_{ji}}{1-(1+\beta)p_{ji}} \right\}$, where $\beta \in\left(0, \frac{1}{\o{p}}-1\right)$ is a positive constant. Set $\eta=\frac{a}{\sqrt{T}}$ for positive constant $a$. Choose $\delta \in \left[\frac{1-\sqrt{1-8\eta^2\omega}}{4 \eta^{2}}, \frac{1+\sqrt{1- 8\eta^2\omega}}{4 \eta^{2}}\right]$. Set $\zeta=\frac{1}{T}$ and $\alpha=\frac{1}{rT}$. Then, under Assumptions \ref{assump_X}-\ref{assump_int}, for $T> \max\{8a^2\omega, \frac{1}{r}\}$, the regret of Problem \eqref{Pro_def} is upper bounded by:
	\begin{align*}  
		&\sum_{t= 1}^{T} \left(\b{E}\left[f^{(t)}\left(\bf{x}^{(t)}\right)\right] - f^{(t)}\left(\bf{x}^*\right)\right) \notag\\
		&\leq \left(\frac{2R^2}{a} + amC^2 + \frac{2mn\w{G}RC}{ar\delta} \right)\sqrt{T} + 2nG\left(1 + \frac{R}{r} \right)\notag\\
		&+ \frac{a}{2}d^2G^2\!\left(2n +\! \sum_{i=1}^n \sum_{j\in\ca{N}_{i}}\! \frac{5\left(1\!+\!\frac{1}{\beta}\right)p_{ij}}{1-(1+\beta)p_{ij}} \!\right)\!\!\sqrt{T} = \ca{O}\left(\sqrt{T}\right).
	\end{align*}
	Further, for any $(i, j)\in\ca{E}$, the constraint violation is upper bounded by:
	\begin{align*}
		&\b{E}\left[\sum_{t=1}^{T} g_{i j}\left(\bf{x}_{i}^{(t)}, \bf{x}_{j}^{(t)}\right)\right] \\
		&\leq  2\sqrt{nG} \left(\frac{1}{a} + a\delta + 2a\w{G}^2\right)^{\frac{1}{2}} \left(T^{\frac{3}{4}}\sqrt{R} + T^{\frac{1}{4}}\sqrt{1+\frac{R}{r} }\right)\\
		&+ \left(1+ a^2\delta + 2a^2\w{G}^2\right)^{\frac{1}{2}} \left( \frac{4R^2}{a^2} + 2mC^2 + \frac{4mn\w{G}RC}{a^2r\delta} \right.\\
		&\left. + d^2G^2\left(2n+\sum_{i=1}^n \sum_{j\in\ca{N}_{i}}\frac{5\left(1+\frac{1}{\beta}\right)p_{ij}}{1-(1+\beta)p_{ij}}\right)\right)^{\frac{1}{2}} \!\!\!\sqrt{T} = \ca{O}\left(T^{\frac{3}{4}}\right)  . 
	\end{align*}
\end{thm}
\begin{rem} \label{rem_BF}
	 Theorem \ref{t2} reveals that, with only two-point bandit feedback, Algorithm \ref{alg_BF} is robust against random link failures with heterogeneous failure probabilities, and maintains the same orders of regret $\ca{O}(\sqrt{T})$ and constraint violations $\ca{O}(T^\frac{3}{4})$ as the full information scenario. This can be explained as follows. First, for any $i\in[n]$ and $t\in[T]$, the estimated gradient $\frac{d}{2 \zeta}\left[f_i^{(t)}\left(\bf{x}_{i} +\zeta \bf{u}_{i}\right) - f^{(t)}_i\left(\bf{x}_{i}-\zeta \bf{u}_{i}\right)\right]\bf{u}_{i}$ based on the values of $f_i^{(t)}$ at the two points $\bf{x}_{i} \pm \zeta \bf{u}_{i}$ is conditionally unbiased and able to track the true gradient $\nabla \w{f}_{i}^{(t)}(\bf{x}_{i})$ of $\w{f}^{(t)}_i(\mathbf{x}_i)$ in expectation. Then, the alternating primal and dual gradient updates in Algorithm \ref{alg_BF} lead us to the saddle-point of the Lagrangian in \eqref{Equ_band_Lag} in expectation. Referring to the result of Theorem \ref{t1}, we know that $\ca{O}(\sqrt{T})$ regret and $\ca{O}(T^\frac{3}{4})$ constraint violation for Problem \eqref{Pro_def} with the cost function replaced by $\sum_{t=1}^T\sum_{i=1}^n\w{f}_i^{(t)}(\bf{x}_i^{(t)})$ are guaranteed. Second, during our analysis, we bound the gap between the cost functions $\sum_{t=1}^T\w{f}^{(t)}(\bf{x}^{(t)})$ and $\sum_{t=1}^Tf^{(t)}(\bf{x}^{(t)})$ by a constant, and bound the term related to the accumulated constraint perturbations by $\ca{O}(\sqrt{T})$ (cf. Section \ref{Sec_BF_Analysis}). Neither affects the orders of both the regret and the constraint violations. Therefore, we have the result in Theorem \ref{t2}.     	
\end{rem}
\begin{rem}
	By Theorem \ref{t2}, the regret bound in the bandit feedback scenario grows quadratically w.r.t. the dimension of agent's decision $d$. The higher the decision dimension, the larger the regret becomes. Since the parameter $d$ is introduced to the regret bound via the gradient estimation based on pointwise values, this demonstrates a performance degradation due to the bandit feedback.
\end{rem}


\subsection{Performance Analysis} \label{Sec_BF_Analysis}

Next, we proceed to prove Theorem \ref{t2}.
\begin{lem}  \label{Lem_band_L-L}
	Consider the update steps in Algorithm \ref{alg_BF}. Under Assumptions \ref{assump_X}-\ref{assump_int}, for any $\bf{x} \in \ca{X}^{n}$, $\@{\lambda} \in \b{R}_+^m$, and $\forall t\in[T]$, the Lagrangian in \eqref{Equ_band_Lag} satisfies:
	\begin{align} \label{Eq_Lem_4}
		&\w{\ca{L}}^{(t)}\left(\bf{x}^{(t)},\@{\lambda}\right) - \w{\ca{L}}^{(t)}(\bf{x},\@\lambda^{(t)}) \notag\\
		&\leq \frac{1}{2 \eta}\left(\left\|\@{\lambda} - \@{\lambda}^{(t)}\right\|^{2} - \left\|\@{\lambda} - \@{\lambda}^{(t+1)}\right\|^{2}\right) + \frac{\eta}{2}\left\|\bf{r}^{(t)}\right\|^{2} \notag\\
		&~~~+ \frac{1}{2 \eta}\left(\left\|\bf{x} - \bf{x}^{(t)}\right\|^{2} - \left\|\bf{x} - \bf{x}^{(t+1)}\right\|^{2}\right) + \frac{\eta}{2}\left\|\w{\bf{q}}^{(t)}\right\|^{2} \notag \\
		&~~~+ \left\langle \@{\lambda} - \@{\lambda}^{(t)}, \nabla_{\@{\lambda}} \w{\ca{L}}^{(t)}\left(\bf{x}^{(t)},  \@{\lambda}^{(t)}\right) - \bf{r}^{(t)}\right\rangle \notag \\
		&~~~+ \left\langle \bf{x}^{(t)} - \bf{x}, \nabla_{\bf{x}} \w{\ca{L}}^{(t)}\left(\bf{x}^{(t)}, \@{\lambda}^{(t)}\right) - \w{\bf{q}}^{(t)}\right\rangle.
	\end{align}
\end{lem}
The proof of Lemma \ref{Lem_band_L-L} is presented in Section H of the supplementary file. The last two terms in \eqref{Eq_Lem_4} can be bounded as follows.
\begin{lem}  \label{Lem_band_grad_gap}
	For any $t\in[T]$, we have
	\begin{align*}
		1)~& \b{E}\left\langle \@{\lambda} - \@{\lambda}^{(t)}, \nabla_{\@{\lambda}} \w{\ca{L}}^{(t)}\left(\bf{x}^{(t)},  \@{\lambda}^{(t)}\right) - \bf{r}^{(t)}\right\rangle \\
		&\leq \eta\w{G}^2 \left( \left\|\@{\lambda}\right\|^2 + \b{E}\left\|\@{\lambda}^{(t)}\right\|^2 \right) + \frac{1}{4\eta} \sum_{i=1}^n\sum_{j\in\ca{N}_i} \b{E}\left[\epsilon_{ij}^{(t)}\right] ;\\
		2)~& \b{E}\left\langle \bf{x}^{(t)} - \bf{x}, \nabla_{\bf{x}} \w{\ca{L}}^{(t)}\left(\bf{x}^{(t)}, \@{\lambda}^{(t)}\right) - \w{\bf{q}}^{(t)}\right\rangle \\
		&\leq 4R^2L^2\eta \b{E}\left\|\@{\lambda}^{(t)}\right\|^{2} + \frac{1}{\eta} \sum_{i=1}^n \sum_{j \in \ca{N}_{i}} \b{E}\left[\epsilon_{ij}^{(t)}\right] .
	\end{align*}
\end{lem}
The proof of Lemma \ref{Lem_band_grad_gap} is presented in Section I of the supplementary file. The squares of $l_2$ norms of the approximated gradients $\w{\bf{q}}^{(t)}$ and $\bf{r}^{(t)}$ for any $t\in[T]$ are bounded in the following lemma.
\begin{lem} \label{Prop_band_grad_bound}
	For any $t\in[T]$, the approximated primal and dual gradients used in Algorithm \ref{alg_BF} satisfy:
	\begin{enumerate} [label=(\alph*) ]
		\item $\left\|\w{\bf{q}}_i^{(t)} \right\|^{2} \leq 2d^2G^2 + 8|\ca{N}_i|\w{G}^2\left\|\@{\lambda}_i^{(t)}\right\|^{2}$, $\forall i\in[n]$;
		\item $\left\|\w{\bf{q}}^{(t)} \right\|^{2} \leq 2nd^2G^2 + 4m\w{G}^2\left\|\@{\lambda}^{(t)}\right\|^{2}$;
		\item $\left\|\bf{r}^{(t)} \right\|^{2} \leq 2mC^2 +2 \delta^{2} \eta^{2} \left\|\@{\lambda}^{(t)}\right\|^{2}$ ;
	\end{enumerate}
	where $d$ is the dimension of agent decisions $\bf{x}_i$ for any $i\in[n]$.  
\end{lem}  
The proof of Lemma \ref{Prop_band_grad_bound} is presented in Section J of the supplementary file. With the results of Lemmas \ref{Lem_band_L-L}-\ref{Prop_band_grad_bound}, we have the following Lemma \ref{Lem_band_Lag_sum}.
\begin{lem}  \label{Lem_band_Lag_sum}
	For any $\bf{x} \in \w{\ca{X}}^{n}$ and $\@{\lambda} \in \b{R}_+^m$, the sum of the Lagrangian in \eqref{Equ_band_Lag} over $T$ iterations satisfies:
	\begin{align}  \label{Equ_band_Lag_sum}
		&\sum_{t= 1}^{T}\b{E}\left[\w{\ca{L}}^{(t)}\left(\bf{x}^{(t)}, \@{\lambda}\right) - \w{\ca{L}}^{(t)}\left((1-\alpha)\bf{x}^* + \alpha \o{\bf{z}}_0,\@\lambda^{(t)}\right)\right] \notag\\
		&\leq \left(\frac{1}{2 \eta} + \eta T\w{G}^2\right) \left\|\@{\lambda}\right\|^2  +  \frac{2R^2}{\eta} + \eta T\left(mC^2 + nd^2G^2 \right) \notag\\
		&~~~+ \frac{\eta}{2}\left(2\delta^{2} \eta^{2} + (2+4m)\w{G}^2 + 8R^2L^2 \right)\sum_{t= 1}^{T} \b{E}\left\|\@{\lambda}^{(t)}\right\|^{2} 	 \notag\\
		&~~~+ \frac{5}{4\eta}\sum_{t= 1}^{T} \sum_{i=1}^n \sum_{j \in \ca{N}_{i}} \b{E}\left[\epsilon_{ij}^{(t)} \right], 
	\end{align}
	where $\o{\bf{z}}_0:=[\bf{z}_0^\s{T},\cdots,\bf{z}_0^\s{T}]^\s{T}\in\b{R}^{nd}$.
\end{lem}
The proof of Lemma \ref{Lem_band_Lag_sum} is presented in Section K of the supplementary file.

\begin{lem}  \label{Prop_band_error}
	The sum of expected error $\b{E}\left[\epsilon_{ij}^{(t)} \right]$ over $i\in[n],j\in\ca{N}_i$, and $t\in[T]$ satisfies:
	\begin{align*}
		\sum_{t=1}^T\sum_{i=1}^n \sum_{j \in \ca{N}_{i}} \b{E}\left[\epsilon_{ij}^{(t)} \right] 
		&\leq \frac{2}{5}\gamma \eta^2\w{G}^2 \sum_{t=1}^T \b{E}\left\|\@{\lambda}^{(t)}\right\|^{2} \\
		&~~+ \sum_{i=1}^n \sum_{j \in \ca{N}_{i}}\frac{2\left(1+\frac{1}{\beta}\right)p_{ij}}{1-(1+\beta)p_{ij}}\eta^2Td^2G^2,
	\end{align*}
	where $\gamma := \max_{i} \left\{\sum_{j\in\ca{N}_{i}} \frac{20\left(1+\frac{1}{\beta}\right)p_{ji}}{1-(1+\beta)p_{ji}} \right\}$ and $\beta \in\left(0, \frac{1}{\o{p}}-1\right)$ is a positive constant. 
\end{lem}
The proof of Lemma \ref{Prop_band_error} is presented in Section L of the supplementary file.

\textbf{Proof of Theorem \ref{t2}:} Plugging the bound in Lemma \ref{Prop_band_error} into \eqref{Equ_band_Lag_sum} yields 
\begin{align}  \label{Eq_band_bound}
	&\sum_{t= 1}^{T} \left(\b{E}\left[\w{f}^{(t)}\left(\bf{x}^{(t)}\right)\right] - \w{f}^{(t)}\left((1-\alpha)\bf{x}^* + \alpha \bf{z}_0\right)\right) \notag\\
	&~~~+ \sum_{t=1}^{T}\b{E}\left\langle\@{\lambda}, \bf{g}\left(\bf{x}^{(t)}\right)\right\rangle - \frac{1}{2} \left(\frac{1}{\eta} + \eta T\left(\delta + 2\w{G}^2\right)\right) \|\@{\lambda}\|^{2} \notag\\
	&~~~- \sum_{t = 1}^{T} \b{E}\left\langle\@{\lambda}^{(t)}, \bf{g}\left((1-\alpha)\bf{x}^* + \alpha \bf{z}_0\right)\right\rangle  \notag\\
	&\leq \frac{\eta}{2} \left(2\delta^{2} \eta^{2} - \delta + \omega \right)\sum_{t= 1}^{T} \b{E}\left\|\@{\lambda}^{(t)}\right\|^{2} + \frac{2R^2}{\eta} + \eta TmC^2 \notag\\
	&~~~+ \frac{\eta T}{2}d^2G^2\left( 2n+ \sum_{i=1}^n \sum_{j\in\ca{N}_{i}}\!\!\frac{5\left(1+\frac{1}{\beta}\right)p_{ij}}{1-(1+\beta)p_{ij}} \!\right)\!, 
\end{align}	
where $\omega := \left(2+4m+ \gamma\right)\w{G}^2 + 8R^2L^2$. Similarly, we choose $\eta = \frac{a}{\sqrt{T}}$ and $\delta \in\left[\frac{1-\sqrt{1-8\eta^2\omega}}{4 \eta^{2}}, \frac{1+\sqrt{1- 8\eta^2\omega}}{4 \eta^{2}}\right]$.
Then, the first term on the right side of \eqref{Eq_band_bound} is non-positive and can be omitted. 

Next, we connect \eqref{Eq_band_bound} with the regret bound of problem \ref{Pro_def}. First, for any $i\in[n]$, we have 
\begin{align} \label{Eq_f_gap}
	&\left|\w{f}^{(t)}\left(\bf{x}^{(t)}\right) - f^{(t)}\left(\bf{x}^{(t)}\right)\right| \notag\\
	&\aleq \sum_{i=1}^n\b{E}_{\bf{v}_i^{(t)}\sim\ca{U}(\b{B})}\left|f_{i}^{(t)}\left(\bf{x}_i^{(t)}+\zeta\bf{v}_i^{(t)}\right)-f_{i}^{(t)}\left(\bf{x}_i^{(t)}\right)\right| \notag\\
	&\bleq \sum_{i=1}^n\b{E}_{\bf{v}_i^{(t)}\sim\ca{U}(\b{B})} G\left\|\zeta\bf{v}_i^{(t)}\right\| \leq nG\zeta,
\end{align}
where $(a)$ is based on the convexity of $l_1$ norm and $(b)$ is from the $G$-Lipschitz continuity of $f_{i}^{(t)}(\cdot)$, for any $i\in[n]$ and $t\in[T]$. Thus, 
\begin{align}
	&\w{f}^{(t)}\left(\bf{x}^{(t)}\right) \geq f^{(t)}\left(\bf{x}^{(t)}\right) - nG\zeta. \label{t2_11}
\end{align}
In addition, we have 
\begin{align*}
	&\left|\w{f}^{(t)}((1-\alpha)\bf{x}^*+\alpha \o{\bf{z}}_0) - f^{(t)}(\bf{x}^*)\right|\\
	&\leq\sum_{i=1}^n\left|\w{f}_{i}^{(t)}((1-\alpha)\bf{x}_i^*+\alpha\bf{z}_0)-f_{i}^{(t)}((1-\alpha)\bf{x}_i^*+\alpha\bf{z}_0)\right|\\
	&~~~+\sum_{i=1}^n\left|f_{i}^{(t)}((1-\alpha)\bf{x}_i^*+\alpha\bf{z}_0)-f_{i}^{(t)}(\bf{x}_i^*)\right|\\
	&\aleq nG\zeta + \sum_{i=1}^nG\|-\alpha\bf{x}_i^*+\alpha\bf{z}_0\| \\
	&\bleq nG\zeta + 2nGR\alpha,
\end{align*}
where $(a)$ is from \eqref{Eq_f_gap} and the $G$-Lipschitz continuity of $f_{i}^{(t)}(\cdot)$, $\forall i\in[n],t\in[T]$, and $(b)$ is based on the boundedness of the set $\ca{X}$. Then, we have
\begin{align}
	\w{f}^{(t)}((1-\alpha)\bf{x}^*+\alpha \o{\bf{z}}_0) \leq f^{(t)}(\bf{x}^*) + nG\zeta + 2nGR\alpha.\label{t2_12}
\end{align}
Subtracting \eqref{t2_12} from \eqref{t2_11} gives
\begin{align}
	f^{(t)}\left(\bf{x}^{(t)}\right) - f^{(t)}(\bf{x}^*) \leq& \w{f}^{(t)}\left(\bf{x}^{(t)}\right) - \w{f}^{(t)}((1-\alpha)\bf{x}^*+\alpha \o{\bf{z}}_0) \notag\\
	&+  2nG\zeta + 2nGR\alpha. \label{t2_13}
\end{align}
Further, from the $\w{G}$-Lipschitz continuity of $g_{ij}$, $\forall (i, j)\in\ca{E}$, we have
\begin{align*}
	\left|g_{ij}\left((1-\alpha)\bf{x}^*+ \alpha \o{\bf{z}}_0 \right) - g_{ij}(\bf{x}^*)\right|\leq \w{G}\alpha \|\bf{x}^*- \o{\bf{z}}_0\|\leq 2n\w{G}R\alpha.
\end{align*}
This gives us the following inequality:
\begin{align*}
	&\left\langle\@{\lambda}^{(t)}, \bf{g}\left((1-\alpha) \bf{x}^{*}+\alpha \w{\bf{y}}_{0}\right)\right\rangle \\
	&=\left\langle\@{\lambda}^{(t)}, \bf{g}\left(\bf{x}^{*}\right)\right\rangle + \left\langle\@{\lambda}^{(t)}, \bf{g}\left((1-\alpha) \bf{x}^{*}+\alpha \w{\bf{y}}_{0}\right)-\bf{g}\left(\bf{x}^{*}\right)\!\right\rangle \\
	& \aleq  \sum_{i=1}^n\sum_{j\in\ca{N}_i} \lambda_{ij}^{(t)} \left|g_{ij}\left((1-\alpha)\bf{x}^*+ \alpha \o{\bf{z}}_0 \right) - g_{ij}(\bf{x}^*)\right| \\ 
	& \leq 2n\w{G}R\alpha \left\|\@{\lambda}^{(t)}\right\|_1 ,
\end{align*}
where $(a)$ is from the fact that $\sum_{t=1}^{T}\left\langle\@{\lambda}^{(t)}, \bf{g}\left(\bf{x}^{*}\right)\right\rangle \leq 0$ and Cauchy-Schwarz inequality. 
Based on the gradient update step of $\@\lambda^{(t)}$, we have
\begin{align*}
	\left\|\@{\lambda}^{(t+1)}\right\|_1 &= \left\|(1-\delta\eta^2)\@{\lambda}^{(t)} + \eta \bf{g}\left(\bf{x}^{(t)}\right) \right\|_1 \\
	&\leq (1-\delta\eta^2)\left\|\@{\lambda}^{(t)}\right\|_1 + \eta mC \\
	&= \sum_{k=0}^{t-1}(1-\delta\eta^2)^k \eta mC.
\end{align*}
Then, the sum of $\left\|\@{\lambda}^{(t+1)}\right\|_1$ over $t\in[T]$ satisfies:
\begin{align*}
	\sum_{t=1}^{T}\left\|\@{\lambda}^{(t)}\right\|_1 &= \sum_{t=1}^{T}\sum_{k=0}^{t-2}(1-\delta\eta^2)^k \eta mC \\
	&= \sum_{k=1}^{T-1}\sum_{t=k+1}^{T}(1-\delta\eta^2)^{t-k-1} \eta mC \\
    & \leq \frac{TmC}{\delta\eta}.
\end{align*}
Thus, we have
\begin{align} \label{Eq_g_gap}
	\sum_{t=1}^{T}\left\langle\@{\lambda}^{(t)}, \bf{g}\left((1-\alpha) \bf{x}^{*}+\alpha \w{\bf{y}}_{0}\right)\right\rangle
	\leq 2mn\w{G}RC \frac{\alpha T}{\delta\eta}.
\end{align}
Plugging the results in \eqref{t2_13} and \eqref{Eq_g_gap} into \eqref{Eq_band_bound}, we have
\begin{align}  \label{Eq_band_bound2}
	&\sum_{t= 1}^{T} \left(\b{E}\left[f^{(t)}\left(\bf{x}^{(t)}\right)\right] - f^{(t)}\left(\bf{x}^*\right)\right) \notag\\
	&\quad+ \sum_{t=1}^{T}\b{E}\left\langle\@{\lambda}, \bf{g}\left(\bf{x}^{(t)}\right)\right\rangle - \frac{1}{2} \left(\frac{1}{\eta} + \eta T\left(\delta + 2\w{G}^2\right)\right) \|\@{\lambda}\|^{2} \notag\\
	&\leq \frac{2R^2}{\eta} + \eta TmC^2 +  2mn\w{G}RC \frac{\alpha T}{\delta\eta} + 2nG\zeta T + 2nGR\alpha T \notag\\
	&\quad+\frac{\eta T}{2}d^2G^2 \left( 2n +  \sum_{i=1}^n \sum_{j\in\ca{N}_{i}} \frac{5\left(1 + \frac{1}{\beta}\right)p_{ij}}{1-(1+\beta)p_{ij}} \right).
\end{align}	
In \eqref{Eq_band_bound2}, three extra terms are introduced due to the bandit feedback, namely, $2mn\w{G}RC \frac{\alpha T}{\delta\eta} $, $ 2nG\zeta T$ and $2nGR\alpha T$. Following the full information scenario, we choose $\@{\lambda}=\frac{\left[\b{E}\left[\sum_{t=1}^{T} \bf{g}\left(\bf{x}^{(t)}\right)\right]\right]^+}{\frac{1}{\eta} + \eta T\left(\delta + 2\w{G}^2\right)}$ and set $\eta=\frac{a}{\sqrt{T}}$. Then, to ensure that the bandit feedback does not degrade the order of convergence bound, we require $\alpha \leq \ca{O}(\frac{1}{T})$. As $\alpha$ is chosen within $\left[\frac{\zeta}{r},1\right)$, this gives $\zeta \propto  \alpha$. Thus, we set $\eta=\frac{a}{\sqrt{T}}$, $\zeta=\frac{1}{T}$, and $\alpha=\frac{1}{rT}$. Plugging the above settings into \eqref{Eq_band_bound2}, the regret bound in Theorem \ref{t2} is obtained. Further, by applying the bound in \eqref{Eq_OptGap} into \eqref{Eq_band_bound2}, we have
\begin{align*}  
	&\sum_{i=1}^{n} \sum_{j \in \ca{N}_{i}} \frac{\left(\left[\b{E}\left[\sum_{t=1}^{T} g_{i j}\left(\bf{x}_{i}^{(t)}, \bf{x}_{j}^{(t)}\right)\right]\right]^{+}\right)^{2}}{2\left(\frac{1}{\eta} + \eta T\left(\delta + 2\w{G}^2\right)\right)} \notag\\
	&\leq 2nGRT +  2nG\left(\!1\!+\!\frac{R}{r}\right) \!+\!\left(\!\frac{2R^2}{a} \!+ amC^2  + \frac{2mn\w{G}RC}{ar\delta} \right.\\
	&~~ + \left.ad^2G^2\!\left( n + \sum_{i=1}^n \sum_{j\in\ca{N}_{i}} \frac{5\left(1+\frac{1}{\beta}\right)p_{ij}}{2(1-(1+\beta)p_{ij})} \right)\right)\!\sqrt{T}.
\end{align*}
Applying the above bound for each given $\left(\left[\b{E}\left[\sum_{t=1}^{T} g_{i j}\left(\bf{x}_{i}^{(t)}, \bf{x}_{j}^{(t)}\right)\right]\right]^{+}\right)^{2}$ and $(i, j)\in\ca{E}$, and taking the square root on both sides, the constraint violations in Theorem \ref{t2} is derived.

\section{Numerical Results}
\label{Results}
In this section, we corroborate the effectiveness of our algorithms and analytical results through numerical simulations. We consider two examples, namely, decentralized online quadratically constrained quadratic program (QCQP) and decentralized online logistic regression. 

\subsection{Decentralized Online QCQP}

\begin{figure}
	\renewcommand\figurename{\small Fig.}
	\centering \vspace*{8pt} \setlength{\baselineskip}{10pt}
	\subfigure[Relative time-average regret]{
		\includegraphics[scale = 0.63]{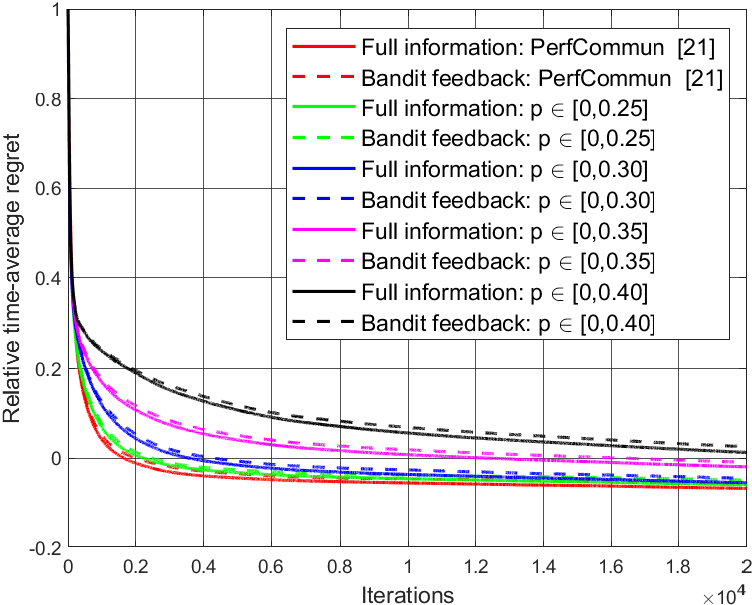}}
	\subfigure[Relative time-average constraint violation]{
		\includegraphics[scale = 0.63]{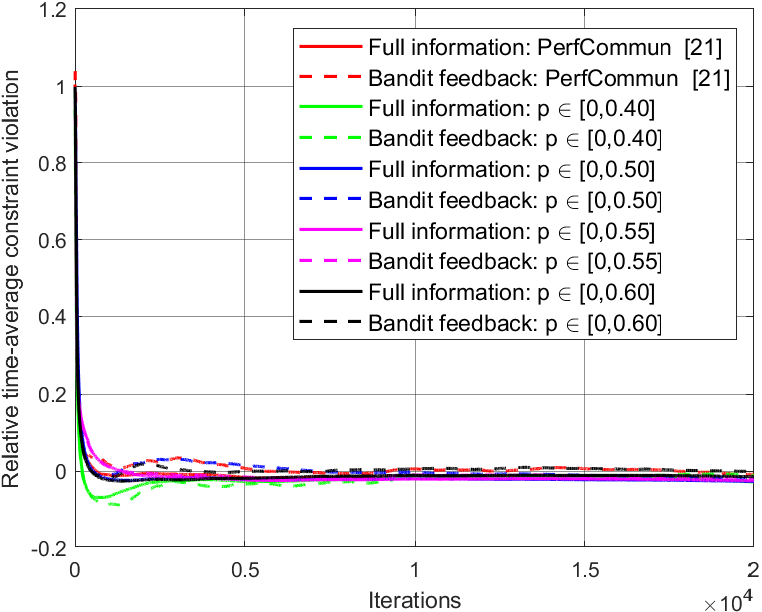}}	
	\caption{Distributed online QCQP}\label{simu_qcqp}
\end{figure}

We first consider decentralized online QCQP over a network of $n$ agents. For each agent $i\in[n]$, the initial cost function is $f_{i}^{(1)}(\bf{x}_i)=\frac{1}{2}\bf{x}_i^\s{T}\bf{A}_i^{(1)}\bf{x}_i+ \bf{x}_i^\s{T}\bf{b}_i^{(1)}$, where $\bf{A}_i^{(1)} \in \b{R}^{d\times d}$ is sampled from a Wishart distribution with $d$ degrees of freedom, and the entries of $\bf{b}_i^{(1)}$ obey a Gaussian distribution with mean and variance drawn uniformly from the interval $[0,1]$. For any $i\in[n]$ and $t\in[T]$, given $f_{i}^{(t)}(\cdot)$, the next cost function $f_{i}^{(t+1)}(\cdot)$ is generated as follows. First, we generate two perturbation matrices $\@\Delta_{\bf{A}_i}^{(t)} \in \b{R}^{d\times d}$ and $\@\Delta_{\bf{b}_i}^{(t)}\in \b{R}^{d}$ with entries drawn uniformly from $[-0.01,0.01]$. Then, we conduct the eigendecomposition for $\bf{A}_i^{(t)} + \@\Delta_{\bf{A}_i}^{(t)}$ and project all of its eigenvalues to $[0,10]$ to obtain $\bf{A}_i^{(t+1)}$. Similarly, $\bf{b}_i^{(t+1)}$ is obtained by projecting the entries of $\bf{b}_i^{(t)} + \@\Delta_{\bf{b}_i}^{(t)}$ into $[-10,10]$. The constraint function associated with agent $i$ and agent $j$ is $ g_{i j}\left(\bf{x}_{i}, \bf{x}_{j}\right)= \frac{1}{2}\left[\bf{x}_{i}^\s{T}, \bf{x}_{j}^\s{T}\right] \bf{S}_{i j}\left[\bf{x}_{i}^\s{T}, \bf{x}_{j}^\s{T}\right]^\s{T}+ \left[\bf{x}_{i}^\s{T}, \bf{x}_{j}^\s{T}\right]\bf{h}_{i j}^\s{T}  + q_{i j} $, $\forall i\in[n], j\in\ca{N}_i$, where the parameters $\left\{\bf{S}_{i j}, \bf{h}_{i j}, q_{i j}\right\} $ are generated in a symmetric manner so that $ g_{i j}\left(\bf{x}_{i}, \bf{x}_{j}\right)=g_{j i}\left(\bf{x}_{j}, \bf{x}_{i}\right) $. 
The objective of the overall network is to solve the following online QCQP:
\begin{align*}
	\underset{\bf{x}_i^{(t)} \in \ca{X}, \forall i,t}{\operatorname{min}} ~&\sum_{t=1}^{T} \sum_{i=1}^{n} \left[\frac{1}{2}\left(\bf{x}_{i}^{(t)}\right)^\s{T} \bf{A}_{i}^{(t)} \bf{x}_{i}^{(t)} + \left(\bf{x}_{i}^{(t)}\right)^\s{T}\bf{b}_{i}^{(t)} \right] \\
	\text{subject to}  ~& \sum_{t=1}^{T}\!\left[\frac{1}{2}\left(\bf{x}_{i}^{(t)}\right)^\s{T}\!\!, \left(\bf{x}_{j}^{(t)}\right)^\s{T}\right]\!\bf{S}_{i j}\!\left[\left(\bf{x}_{i}^{(t)}\right)^\s{T}\!\!, \left(\bf{x}_{j}^{(t)}\right)^\s{T}\right]^\s{T} \notag\\
	&+\left[\left(\bf{x}_{i}^{(t)}\right)^\s{T}\!,\! \left(\bf{x}_{j}^{(t)}\right)^\s{T}\right]\bf{h}_{i j} + q_{i j} \leq 0 ,
	\forall (i, j)\in\ca{E}. 
\end{align*}

We conduct our experiment over a randomly generated Erdos-Renyi graph with $n=30$ agents. The probability of an edge between any pair of nodes of the Erdos-Renyi graph is $0.2$. The data dimension is $d=3$. The feasible set $\ca{X}$ is a Euclidean ball centered at the origin with radius $R = \sqrt{d}$. The decision of agent $i$ is initialized as $\bf{x}_i^{(1)} \sim\Pi_{\ca{X}}(\ca{N}(\bf{0},\bf{I}))$, $\forall i\in[n]$. We compare our algorithm with the perfect communication case \cite{koppel2017proximity}, which is termed by ``Perfcommun [21]'' in the legends of all figures. In addition, we consider four different link failure cases that $p_{ij}$ uniformly drawn from $[0,0.25]$, $[0,0.30]$, $[0,0.35]$, and $[0,0.40]$ for any $(i, j)\in\ca{E}$. For notational convenience, we represent ``$p_{ij}$ $\forall (i, j)\in\ca{E}$'' by $p$ in the legends of all figures. The stepsize is $\eta = \frac{0.12}{\sqrt{T}}$ and the regularization parameter is $\delta=1$. The parameter $r$ in the bandit feedback scenario is set to be $\frac{R}{2}$.

In the simulation, the relative time-average regret $\frac{\text{Reg}(t)}{t\cdot\text{Reg}(1)}$ and the relative time-average constraint violation $\frac{\text{Vio}_{(ij)}(t)}{t\cdot\text{Vio}_{(ij)}(1)}$ are used as performance metrics, which are shown in Fig.~\ref{simu_qcqp}-(a) and Fig.~\ref{simu_qcqp}-(b), respectively. From Fig.~\ref{simu_qcqp}, we observe that for all the considered link failure cases, both the time-average regret and the time-average constraint violation converge to zero or a number slightly below than zero as time goes to infinity. This corroborates the sublinearity of the regret and the constraint violations, as shown in Theorems \ref{t1} and \ref{t2}. Moreover, Fig.~\ref{simu_qcqp} shows that the greater the probability of link failures, the larger the regret becomes. This is in accordance with the regret bounds of Theorems \ref{t1} and \ref{t2}. Further, we see that for any link failure case, the performance of the bandit feedback scenario is close to that of the full information scenario, which verifies the result of Theorem \ref{t2}. Finally, the performance of our algorithm approaches that of the perfect communication case \cite{koppel2017proximity}, validating the effectiveness of our algorithm in combating random link failures.

\subsection{Decentralized Online Logistic Regression}

\begin{figure}
	\renewcommand\figurename{\small Fig.}
	\centering \vspace*{8pt} \setlength{\baselineskip}{10pt}
	\subfigure[Relative time-average regret]{
		\includegraphics[scale = 0.63]{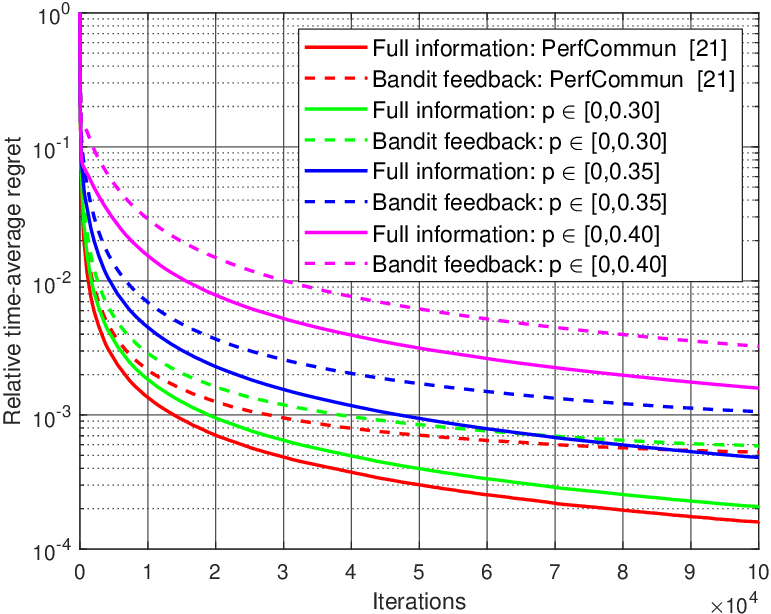}}
	\subfigure[Relative time-average constraint violation]{
		\includegraphics[scale = 0.63]{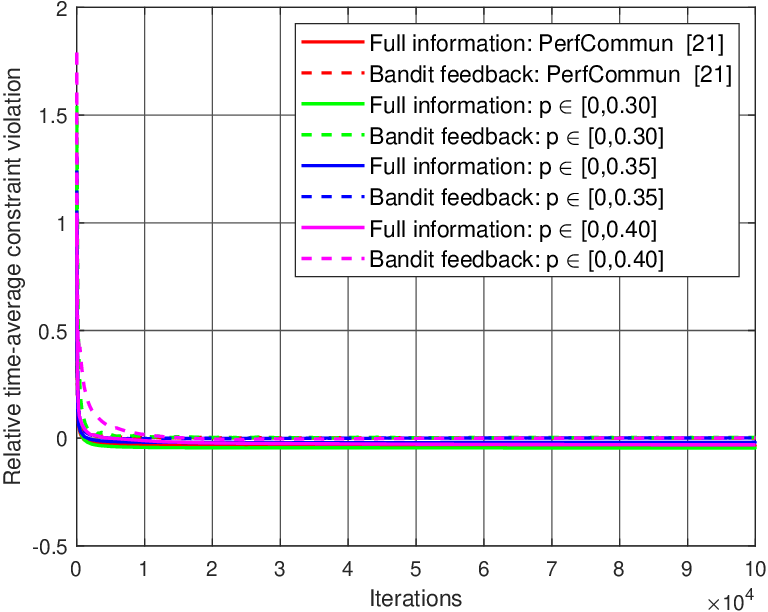}}	
	\caption{Distributed online logistic regression}\label{simu_lr}
\end{figure}

We further study an example of decentralized online logistic regression, where each agent aims to estimate a weight vector (classifier) $ \bf{x}_{i} \in \b{R}^{d} $ associated with the input feature $\@{\psi}_{i} \in \b{R}^{d} $ and the label $\ell_{i} \in\{-1,+1\}$. The logistic regression model assumes that, given an input feature $\@{\psi}_{i}$, the conditional distribution of the label $\ell_{i}$ is $ p\left(\ell_{i} \mid \@{\psi}_{i}; \bf{x}_{i}\right)=\frac{1}{1+e^{- \ell_{i} \@{\psi}_{i}^\s{T} \bf{x}_{i}}}$, $\forall i\in[n]$. The coefficients $\left(\@{\psi}_{i}^{(t)}, \ell_{i}^{(t)}\right)$ are independent across agents, $\forall t\in[T]$. Each pair of neighbors is associated with a proximity constraint on their decisions that $\left\|\bf{x}_{i}-\bf{x}_{j}\right\| \leq b_{ij}$, and we have $b_{ij} = b_{ji}$ for any $(i, j)\in\ca{E}$. The goal of the decentralized online logistic regression is to estimate the weight vectors $\bf{x}$ by solving the following problem:
\begin{align}
	\underset{\bf{x}_i^{(t)} \in \ca{X}, \forall i,t}{\min} ~& \sum_{t=1}^{T} \sum_{i=1}^{n} \left[\log \left(1+e^{-\ell_{i}^{(t)} \left(\@{\psi}_{i}^{(t)}\right)^\s{T} \bf{x}_{i}^{(t)}}\right)\right] \notag\\
	\text{subject to} ~&\frac{1}{T}\sum_{t=1}^{T}\left\|\bf{x}_{i}^{(t)}-\bf{x}_{j}^{(t)}\right\|^{2} \leq b_{i j}^{2}, \forall (i, j)\in\ca{E}.  \label{Eq_LR}
\end{align}

In the simulation, for any $i\in[n]$, the sequential features $\left\{\@{\psi}_{i}^{(t)}\right\}$ are drawn from the distribution $\ca{N}(\bf{0},\bf{I})$. We consider three different link failure cases that $p_{ij}$ uniformly drawn from $[0,0.30]$, $[0,0.35]$, and $[0,0.40]$, for any $(i, j)\in\ca{E}$. The other parameters are set as follows: $n=30$, $\eta=\frac{1}{\sqrt{T}}$, $\delta=1$, $R=1$, and $r=R/2$. The relative time-average regret and the relative time-average constraint violation of decentralized online logistic regression are respectively presented in Fig.~\ref{simu_lr}-(a) and Fig.~\ref{simu_lr}-(b). The results are quantitively analogous to the results of Fig.~\ref{simu_qcqp}. First, the time-average regret and the time-average constraint violation of all link failure cases converge to zero as time goes to infinity, confirming the sublinearity of the regret and the constraint violations. Second, a higher probability of link failures leads to a larger regret. In Fig.~\ref{simu_lr}-(a), the performance gap between the bandit feedback scenario and the corresponding full information scenario results from the gradient approximation error using pointwise values, as discussed in Theorem \ref{t2}.

\section{Conclusions}
\label{Conclusion}
In this paper, we have studied the impact of random link failures on decentralized multi-task OCO with pairwise constraints, where communication between neighbors is unreliable and packet dropping can happen. For the full information scenario, we have developed a robust decentralized saddle-point algorithm against random link failures with heterogeneous probabilities, where a strategy to replace the missing decisions of neighbors with their latest received values was employed. Then, by judiciously bounding the accumulated deviation stemming from this replacement, we have established that our algorithm achieves $\ca{O}(\sqrt{T})$ regret and $\ca{O}(T^\frac{3}{4})$ constraint violations. These two bounds match the performance bounds of the algorithm with perfect communications \cite{koppel2017proximity} in order sense. Further, we have extended our algorithm and analysis to the two-point bandit feedback scenario. Performance bounds of the same orders as the full information scenario have been provided. Finally, the efficacy of the proposed algorithms and the analytical results have been verified through numerical simulations.


\section*{Appendix}

\subsection{Applications of Problem \eqref{Pro_def}}

\textbf{Application 1} (Multi-Robot Path Planning): In decentralized multi-robot path planning,  robots aim to navigate toward their destinations with the minimum path cost. The trajectory of robot $i$ is denoted as $\pi_i$, and its private cost function is represented by $c_i(\cdot)$. The trajectory $\pi_i$ is defined by the displacements of unit-spaced points, given by $\pi_i=\left[\text {start}_i, \mathbf{x}_{i}^{(2)}, \mathbf{x}_{i}^{(3)}, \cdots, \mathbf{x}_{i}^{(T)}, \text {goal}_i\right]^{\top}$, where $\mathbf{x}_{i}^{(t)}$ represents the position of agent $i$ the $t$th time slot, $\forall i\in[n]$ and $t\in[T]$. Given the current position $\mathbf{x}_i^{(t-1)}$, the available search region of $\mathbf{x}_i^{(t)}$ is the unit circle around $\mathbf{x}_i^{(t-1)}$. Moreover, to ensure safe navigation, each pair of neighboring agents $(i,j)\in\ca{E}$ is subject to a distance constraint that
\begin{align*}
    g_{ij}\left(\mathbf{x}_{i},\mathbf{x}_{j}\right) = p_{ij} - \left\|\mathbf{x}_{i} - \mathbf{x}_{j} \right\| \leq 0, \forall (i,j)\in\ca{E}. 
\end{align*}
At each time slot $t\in[T]$, the cost of a trajectory comprises two components: the travel cost, denoted by $\alpha_i \left\|\mathbf{x}_{i} - \mathbf{x}_{i}^{(t-1)} \right\|$, and the penalty for distance to destination, expressed as $\beta_i\left\|\mathbf{x}_{i} - \text {goal}_i\right\|$, which gives the cost function as
\begin{align*}
    c_i^{(t)}(\mathbf{x}_{i}) = \alpha_i \left\|\mathbf{x}_{i} - \mathbf{x}_{i}^{(t-1)} \right\| + \beta_i\left\|\mathbf{x}_{i} - \text {goal}_i\right\|, \forall t\in[T]. 
\end{align*}
Then, the regret and constraint violations of the decentralized multi-robot path planning are respectively given by 
\begin{align*}
    \text{Reg}(T)=&\sum_{t=1}^T\sum_{i=1}^n c_{i}^{(t)}(\mathbf{x}_i^{(t)}) - \sum_{i=1}^n c_i(\pi_i^*),\\
    \text{Vio}_{ij}(T)= & \sum_{t=1}^T \left(p_{ij} - \left\|\mathbf{x}_{i}^{(t)} - \mathbf{x}_{j}^{(t)} \right\|\right), \forall (i,j)\in\ca{E},
\end{align*} 
where  $\pi_i^* = \left[\text {start}_i, \left(\mathbf{x}_{i}^{(2)}\right)^*, \left(\mathbf{x}_{i}^{(3)}\right)^*, \cdots, \left(\mathbf{x}_{i}^{(T)}\right)^*, \text {goal}_i\right]^{\top}$, for all $i\in[n]$ is the ``best'' trajectory, representing a set of positions that collectively minimize the total cost while satisfying the pairwise distance constraints.

\textbf{Application 2} (Correlated Random Field Estimation). In a Gauss-Markov random field, the objective is to estimate the values of the fields at the locations of all sensors, denoted by $\mathbf{x}_i^{*}$, $\forall i \in [n]$, based on a sequence of observations. Let $\boldsymbol{\theta}_{i}^{(t)}$ represent the observation collected by sensor $i$ at time $t$. The observation is the noisy linear transformation of $\mathbf{x}_i$, given by $\boldsymbol{\theta}_{i}^{(t)} = \mathbf{H}_i^{(t)} \mathbf{x}_i + \mathbf{w}_{i}^{(t)}$, where $\mathbf{H}_i^{(t)}$ is the linear transform matrix and $\mathbf{w}_{i}^{(t)} \sim \ca{N}(0, \sigma^2 \mathbf{I})$, $\forall i\in[i], t\in[T]$. The loss function of each agent is the mean square error of the estimation, i.e.,
\begin{align*}
	f_{i}^{(t)}\left(\mathbf{x}_{i}\right)=\left\|\mathbf{H}_i^{(t)} \mathbf{x}_{i}-\boldsymbol{\theta}_{i}^{(t)}\right\|^2.
\end{align*} 
The quality of these estimates can be improved by leveraging correlations between adjacent agents, represented by the constraints 
\begin{align*}
	\left\|\mathbf{x}_{i}^{(t)} - \mathbf{x}_{j}^{(t)}\right\|-\gamma_{i j}\leq 0, \forall (i, j)\in\ca{E}.
\end{align*} 
The network regret and constraint violations are respectively given by
\begin{align*}
    \text{Reg}(T)= & \sum_{t=1}^T \sum_{i=1}^N \left\|\mathbf{H}_i^{(t)} \mathbf{x}_{i}^{(t)}-\boldsymbol{\theta}_{i}^{(t)}\right\|^2 \\
    & - \min_{\mathbf{x}_i,\forall i\in[n]}\sum_{t=1}^T \sum_{i=1}^N\left\|\mathbf{H}_i^{(t)} \mathbf{x}_i-\boldsymbol{\theta}_{i}^{(t)} \right\|^2, \\
    \text{Vio}_{ij}(T)= & \sum_{t=1}^T\left(\left\|\mathbf{x}_{i}^{(t)} - \mathbf{x}_{j}^{(t)}\right\|-\gamma_{i j}\right), \forall (i,j)\in\ca{E} .
\end{align*}
The ``best'' decision here is the sequence of estimations that minimizes the joint mean square errors while optimally utilizing the correlated information.

\subsection{Proof of Proposition \ref{prop}}
 From Assumptions \ref{assump_X} and \ref{assump_g_grad}, $\forall (i, j)\in\ca{E}$ and $\bf{x}_i,\bf{x}_j\in \ca{X}$, we have 
\begin{align*} 
	& \left\|\nabla_{\bf{x}_{i}} g_{i j}\left(\bf{x}_{i}, \bf{x}_{j}\right) - \nabla_{\bf{x}_{i}} g_{i j}\left(\bf{x}_{i}^o, \bf{x}_{j}^o\right)\right\| \\
	& \leq \left\|\nabla_{\bf{x}_{i}} g_{i j}\left(\bf{x}_{i}, \bf{x}_{j}\right) - \nabla_{\bf{x}_{i}} g_{i j}\left(\bf{x}_{i}^o, \bf{x}_{j}\right)\right\| \\
	& \quad + \left\|\nabla_{\bf{x}_{i}} g_{i j}\left(\bf{x}_{i}^o, \bf{x}_{j}\right)- \nabla_{\bf{x}_{i}} g_{i j}\left(\bf{x}_{i}^o, \bf{x}_{j}^o\right)\right\| \\
	& \leq L\left\|\bf{x}_{i} - \bf{x}_{i}^o \right\| + L\left\|\bf{x}_{j} - \bf{x}_{j}^o \right\|\\
	& \leq 4RL.
\end{align*}
Then, we have 
\begin{align*} 
	& \left\|\nabla_{\bf{x}_{i}} g_{i j}\left(\bf{x}_{i}, \bf{x}_{j}\right) \right\| 
	 \leq \left\|\nabla_{\bf{x}_{i}} g_{i j}\left(\bf{x}_{i}^o, \bf{x}_{j}^o\right)\right\| + 4RL.
\end{align*}
From Assumption \ref{assump_g}, $\left\|\nabla_{\bf{x}_{i}} g_{i j}\left(\bf{x}_{i}^o, \bf{x}_{j}^o\right)\right\|$ is bounded. Thus, $\left\|\nabla_{\bf{x}_{i}} g_{i j}\left(\bf{x}_{i}, \bf{x}_{j}\right) \right\|$ is bounded, i.e., there exists a constant $\w{G}$ such that $\left\|\nabla_{\bf{x}_{i}} g_{i j}\left(\bf{x}_{i}, \bf{x}_{j}\right) \right\| \leq \w{G}$. Define $\gamma_i(t) = \bf{x}_{i}^{\prime} + t (\bf{x}_{i} - \bf{x}_{i}^{\prime})$. We have
\begin{align*} 
	&\left|g_{i j}\left(\bf{x}_{i}, \bf{x}_{j}\right) - g_{i j}\left(\bf{x}_{i}^{\prime}, \bf{x}_{j}\right) \right| \\
	& = \left|\int_{0}^{1} \nabla_{\bf{x}_{i}} g_{i j}\left(\gamma_{i}(t), \bf{x}_{j}\right)(\bf{x}_{i} - \bf{x}_{i}^{\prime}) \,dt \right| \\
	& \leq \left|\int_{0}^{1} \left\|\nabla_{\bf{x}_{i}} g_{i j}\left(\gamma_{i}(t), \bf{x}_{j}\right)\right\| \left\|\bf{x}_{i} - \bf{x}_{i}^{\prime}\right\| \,dt \right|\\
	& \leq \w{G} \left\|\bf{x}_{i} - \bf{x}_i^{\prime}\right\|,
\end{align*}
which proves Part $(i)$ of Proposition \ref{prop}.

Similarly, we have 
\begin{align*} 
	&\left|g_{i j}\left(\bf{x}_{i}, \bf{x}_{j}\right) - g_{i j}\left(\bf{x}_{i}^o, \bf{x}_{j}^o\right) \right| \\
	& \leq \left|g_{i j}\left(\bf{x}_{i}, \bf{x}_{j}\right) - g_{i j}\left(\bf{x}_{i}^o, \bf{x}_{j}\right) \right| + \left|g_{i j}\left(\bf{x}_{i}^o, \bf{x}_{j}\right) - g_{i j}\left(\bf{x}_{i}^o, \bf{x}_{j}^o\right) \right| \\
	& \leq \w{G} \left\|\bf{x}_{i} - \bf{x}_{i}^o\right\| + \w{G} \left\|\bf{x}_{j} - \bf{x}_{i}^o\right\| \\
	& \leq 4R\w{G}.
\end{align*}
Then, $ \left|g_{i j}\left(\bf{x}_{i}, \bf{x}_{j}\right) \right| \leq \left|g_{i j}\left(\bf{x}_{i}^o, \bf{x}_{j}^o\right) \right| + 4R\w{G}$.
Thus, there exists a constant $C$ such that $|g_{ij}(\bf{x}_{i}, \bf{x}_{j})| \leq C$, which proves Part $(ii)$ of Proposition \ref{prop}.

\subsection{Proof of Lemma \ref{Lem_L-L}}

By Assumption \ref{assump_convex}, we know that both $f^{(t)}(\bf{x})$ and $\bf{g}(\bf{x})$ are convex in $\bf{x}$, so as the Lagrangian  $\ca{L}^{(t)}\left(\bf{x},\@\lambda\right)$, for any given $\@{\lambda} \in \b{R}_+^m$, $\forall t \in[T]$. Thus, we have 
\begin{align*} 
	&\ca{L}^{(t)}\!\left(\!\bf{x},\@\lambda^{(t)}\!\right)\! \geq \ca{L}^{(t)}\!\left(\!\bf{x}^{(t)},\@\lambda^{(t)}\!\right)\! +\! \nabla_{\bf{x}}\ca{L}^{(t)}\!\left(\!\bf{x}^{(t)},\@\lambda^{(t)}\!\right)^\s{T}\!\!\left(\!\bf{x}-\bf{x}^{(t)}\!\right)\!.
\end{align*}
Further, since $\ca{L}^{(t)}\left(\bf{x}, \@{\lambda}\right)$ is concave in $\@{\lambda}$, we have
\begin{align*} 
	&\ca{L}^{(t)}\!\left(\!\bf{x}^{(t)}, \@{\lambda}\!\right)\! \leq \ca{L}^{(t)}\!\left(\!\bf{x}^{(t)}, \@{\lambda}^{(t)}\!\right)\! +\! \nabla_{\@{\lambda}} \ca{L}^{(t)}\!\left(\!\bf{x}^{(t)},  \@{\lambda}^{(t)}\!\right)^\s{T}\! \!\left(\!\@{\lambda} - \@{\lambda}^{(t)}\!\right)\!.
\end{align*}
Combining the above two inequalities yields
\begin{align}
	&\ca{L}^{(t)}\left(\bf{x}^{(t)},\@{\lambda}\right) - \ca{L}^{(t)}(\bf{x},\@\lambda^{(t)}) \notag\\
	&\leq  \left\langle \@{\lambda} - \@{\lambda}^{(t)}, \nabla_{\@{\lambda}} \ca{L}^{(t)}\left(\bf{x}^{(t)},  \@{\lambda}^{(t)}\right) \right\rangle \notag\\
	&~~~+ \left\langle \bf{x}^{(t)} - \bf{x}, \nabla_{\bf{x}} \ca{L}^{(t)} \left(\bf{x}^{(t)}, \@{\lambda}^{(t)}\right)\right\rangle.  \label{Eq_L-L}
\end{align}
These two terms on the right side of \eqref{Eq_L-L} are handled as follows.

\emph{1)} From the update rule of $\@{\lambda}$ and Fact \ref{Fact_projection}, we have  
\begin{align} \label{Eq_lam_r}
	&\left\|\@{\lambda}-\@{\lambda}^{(t+1)}\right\|^{2} = \left\|\@{\lambda}-\left[\@{\lambda}^{(t)} + \eta \bf{r}^{(t)} \right]^{+}\right\|^{2} \notag\\
	&\leq  \left\|\@{\lambda}-\@{\lambda}^{(t)}\right\|^{2} + \eta^{2} \left\|\bf{r}^{(t)}\right\|^2
	-2 \eta \left\langle \@{\lambda}-\@{\lambda}^{(t)}, \bf{r}^{(t)}\right\rangle. 
\end{align}
Rearranging the terms in \eqref{Eq_lam_r} gives
\begin{align*}  
	&\left\langle \@{\lambda} - \@{\lambda}^{(t)}, \bf{r}^{(t)} \right\rangle \notag\\
	&\leq \frac{1}{2 \eta}\left(\left\|\@{\lambda} - \@{\lambda}^{(t)}\right\|^{2} - \left\|\@{\lambda} - \@{\lambda}^{(t+1)}\right\|^{2}\right) + \frac{\eta}{2}\left\|\bf{r}^{(t)}\right\|^{2}.
\end{align*}
Then, we have
\begin{align*}
	&\left\langle \@{\lambda} - \@{\lambda}^{(t)}, \nabla_{\@{\lambda}} \ca{L}^{(t)}\left(\bf{x}^{(t)},  \@{\lambda}^{(t)}\right) \right\rangle  \\
	&\leq \frac{1}{2 \eta}\left(\left\|\@{\lambda} - \@{\lambda}^{(t)}\right\|^{2} - \left\|\@{\lambda} - \@{\lambda}^{(t+1)}\right\|^{2}\right) + \frac{\eta}{2}\left\|\bf{r}^{(t)}\right\|^{2} \\
	&~~~+ \left\langle \@{\lambda} - \@{\lambda}^{(t)}, \nabla_{\@{\lambda}} \ca{L}^{(t)}\left(\bf{x}^{(t)},  \@{\lambda}^{(t)}\right) - \bf{r}^{(t)}\right\rangle.
\end{align*}

\emph{2)} From the update rule of $\bf{x}$ and Fact \ref{Fact_projection}, we have
\begin{align} \label{Eq_lam_x}
	&\left\|\bf{x}-\bf{x}^{(t+1)}\right\|^{2}  = \left\|\bf{x} - \Pi_{\ca{X}}\left(\bf{x}^{(t)}-\eta \bf{q}^{(t)} \right)\right\|^{2} \notag\\
	&\leq \left\|\bf{x} -\bf{x}^{(t)}\right\|^{2} + \eta^{2}\left\|\bf{q}^{(t)}\right\|^{2} + 2\eta \left\langle \bf{x} -\bf{x}^{(t)}, \bf{q}^{(t)}\right\rangle.
\end{align}
Rearranging the terms in \eqref{Eq_lam_x} gives
\begin{align*} 
	&\left\langle \bf{x}^{(t)} - \bf{x}, \bf{q}^{(t)}\right\rangle \\
	&\leq \frac{1}{2 \eta}\left(\left\|\bf{x} - \bf{x}^{(t)}\right\|^{2} - \left\|\bf{x} - \bf{x}^{(t+1)}\right\|^{2}\right) + \frac{\eta}{2}\left\|\bf{q}^{(t)}\right\|^{2}.
\end{align*}
Then, we have 
\begin{align*}
	&\left\langle \bf{x}^{(t)} - \bf{x}, \nabla_{\bf{x}} \ca{L}^{(t)}\left(\bf{x}^{(t)}, \@{\lambda}^{(t)}\right)\right\rangle \\
	&\leq \frac{1}{2 \eta}\left(\left\|\bf{x} - \bf{x}^{(t)}\right\|^{2} - \left\|\bf{x} - \bf{x}^{(t+1)}\right\|^{2}\right) + \frac{\eta}{2}\left\|\bf{q}^{(t)}\right\|^{2} \\
	&~~~+ \left\langle \bf{x}^{(t)} - \bf{x}, \nabla_{\bf{x}} \ca{L}^{(t)}\left(\bf{x}^{(t)}, \@{\lambda}^{(t)}\right) - \bf{q}^{(t)}\right\rangle.
\end{align*}
Plugging the results in Part \emph{1)} and Part \emph{2)} into \eqref{Eq_L-L} gives Lemma \ref{Lem_L-L}. 

\subsection{Proof of Lemma \ref{Lem_grad_gap}}
\emph{1)} From the definitions of $\frac{\partial}{\partial \lambda_{i j}}\ca{L}^{(t)}(\bf{x}, \@{\lambda})$ in \eqref{dual_grad} and $r_{ij}^{(t)}$ in \eqref{Approx_dual_grad}, for any $(i, j)\in\ca{E}$ and $t\in[T]$, we have
\begin{align*}
	&\left\langle \@{\lambda} - \@{\lambda}^{(t)}, \nabla_{\@{\lambda}} \ca{L}^{(t)}\left(\bf{x}^{(t)},  \@{\lambda}^{(t)}\right) - \bf{r}^{(t)}\right\rangle \notag\\
	&\leq \left\| \@{\lambda} - \@{\lambda}^{(t)} \right\| \left\| \nabla_{\@{\lambda}} \ca{L}^{(t)}\left(\bf{x}^{(t)},\@{\lambda}^{(t)}\right) - \bf{r}^{(t)} \right\| \notag\\
	&\leq \left\| \@{\lambda} - \@{\lambda}^{(t)} \right\| \sum_{i=1}^n\sum_{j\in\ca{N}_i} \left|g_{i j}\left(\bf{x}_{i}^{(t)}, \bf{x}_{j}^{(t)}\right) - g_{i j}\left(\bf{x}_{i}^{(t)}, \bf{x}_{j\to i}^{(t)}\right)\right| \notag\\
	&\aleq \w{G} \left\| \@{\lambda} - \@{\lambda}^{(t)} \right\| \sum_{i=1}^n\sum_{j\in\ca{N}_i}\left\|\bf{x}_{j}^{(t)} - \bf{x}_{j\to i}^{(t)}\right\|  \notag\\
	&\leq \w{G}\left( \left\|\@{\lambda}\right\|+\left\|\@{\lambda}^{(t)}\right\| \right) \sum_{i=1}^n\sum_{j\in\ca{N}_i}\left\|\bf{x}_{j}^{(t)} - \bf{x}_{j\to i}^{(t)}\right\|  \notag\\
	&\leq \eta\w{G}^2 \left( \left\|\@{\lambda}\right\|^2 + \left\|\@{\lambda}^{(t)}\right\|^2 \right) + \frac{1}{4\eta} \sum_{i=1}^n\sum_{j\in\ca{N}_i}\left\|\bf{x}_{j}^{(t)} - \bf{x}_{j\to i}^{(t)}\right\|^2,
\end{align*}
where $(a)$ is from the $\w{G}$-Lipschitz continuity of $g_{ij}(\cdot)$, $\forall (i, j)\in\ca{E}$. 

\emph{2)} From the definitions of $\nabla_{\bf{x}_i} \ca{L}^{(t)}\left(\bf{x}, \@{\lambda}\right)$ in \eqref{primal_grad} and $\bf{q}_i^{(t)}$ in \eqref{Approx_primal_grad}, for any $i\in[n]$ and $t\in[T]$,  we have 
\begin{align*}
	&\left\langle \bf{x}^{(t)} - \bf{x}, \nabla_{\bf{x}} \ca{L}^{(t)}\left(\bf{x}^{(t)}, \@{\lambda}^{(t)}\right) - \bf{q}^{(t)}\right\rangle \\
	&\aeq 2\sum_{i=1}^n\sum_{j \in \ca{N}_{i}}\lambda_{i j}^{(t)} \left\langle \bf{x}_i^{(t)}-\bf{x}_i, \right.\\
	&~~~\left. \nabla_{\bf{x}_{i}} g_{i j}\left(\bf{x}_{i}^{(t)}, \bf{x}_{j}^{(t)}\right) - \nabla_{\bf{x}_{i}} g_{i j}\left(\bf{x}_{i}^{(t)}, \bf{x}_{j\to i}^{(t)}\right)\right\rangle \\
	&\leq 2\sum_{i=1}^n\sum_{j \in \ca{N}_{i}}\lambda_{i j}^{(t)}  \left\|\bf{x}_i^{(t)}-\bf{x}_i\right\| \\
	&~~~ \left\| \nabla_{\bf{x}_{i}} g_{i j}\left(\bf{x}_{i}^{(t)}, \bf{x}_{j}^{(t)}\right) - \nabla_{\bf{x}_{i}} g_{i j}\left(\bf{x}_{i}^{(t)}, \bf{x}_{j\to i}^{(t)}\right) \right\| \\
	&\bleq 4RL\sum_{i=1}^n\sum_{j \in \ca{N}_{i}}\lambda_{i j}^{(t)} \left\|\bf{x}_{j}^{(t)} - \bf{x}_{j\to i}^{t}\right\| \\
	&\leq 4R^2L^2\eta\left\|\@{\lambda}^{(t)}\right\|^{2} + \frac{1}{\eta} \sum_{i=1}^n \sum_{j \in \ca{N}_{i}}\left\|\bf{x}_{j}^{(t)} -\bf{x}_{j\to i}^{(t)} \right\|^2, 
\end{align*}
where (a) is based on the fact that $ \lambda_{i j}^{(t)}=\lambda_{ji}^{(t)} $ in  $\nabla_{\bf{x}} \ca{L}^{(t)}\left(\bf{x}^{(t)}, \@{\lambda}^{(t)}\right)$, $\forall (i, j)\in\ca{E}$, and $t\in[T]$\footnote{ $\nabla_{\bf{x}} \ca{L}^{(t)}\left(\bf{x}^{(t)}, \@{\lambda}^{(t)}\right)$ is the gradient corresponding to the perfect communication case, in which the pairwise dual variables satisfy symmetry that $\lambda_{i j}^{(t)}=\lambda_{ji}^{(t)}, \forall (i, j)\in\ca{E}$, and $t\in[T]$. This can be proved by deduction. First, for $t = 1$, the condition $\lambda_{i j}^{(1)} = \lambda_{ji}^{(1)} = 0$ trivially holds. Suppose that for $t=\tau>1$, $\lambda_{i j}^{(\tau)} = \lambda_{ji}^{(\tau)}$ holds. Then, for $t=\tau+1$, we have $\lambda_{i j}^{(\tau+1)} = \left[\lambda_{i j}^{(\tau)}+\eta\left(g_{i j}\left(\bf{x}_{i}^{(\tau)}, \bf{x}_{j}^{(\tau)}\right)-\delta \eta \lambda_{i j}^{(\tau)}\right)\right]^{+} = \left[\lambda_{ji}^{(\tau)}+\eta\left(g_{ji}\left(\bf{x}_{j}^{(\tau)}, \bf{x}_{i}^{(\tau)}\right)-\delta \eta \lambda_{ji}^{(\tau)}\right)\right]^{+} = \lambda_{ji}^{(\tau+1)}$, which yields the proof.}, and (b) is from the boundedness of set $\ca{X}$ and the $L$-Lipschitz continuity of $\nabla_{\bf{x}_{i}} g_{i j}$, $\forall i\in[n], j\in\ca{N}_i$. By now, Lemma \ref{Lem_grad_gap} is proved. 

\subsection{Proof of Lemma \ref{Prop_grad_bound}} 
We first prove Part \emph{(a)} of Lemma \ref{Prop_grad_bound}. From the definition of $\bf{q}_i^{(t)}$ in \eqref{Approx_primal_grad}, for any $ i\in[n]$ and $t\in[T]$, we have 
\begin{align} \label{Eq_q_i}
	\left\|\bf{q}_i^{(t)} \right\|^{2} &= \left\|\nabla_{\bf{x}_{i}} f_{i}^{(t)}\!\left(\!\bf{x}_{i}^{(t)}\!\right) + 2\!\sum_{j\in\ca{N}_{i}}\!\lambda_{i j}^{(t)} \nabla_{\bf{x}_{i}} g_{i j}\left(\!\bf{x}_{i}^{(t)}\!, \bf{x}_{j\to i}^{(t)}\right)\right\|^{2} \notag\\
	& \leq 2\left\|\nabla_{\bf{x}_{i}} f_{i}^{(t)}\left(\bf{x}_{i}^{(t)}\right)\right\|^{2} \notag\\
	&\quad + 2\left\|2\sum_{j\in\ca{N}_{i}}\lambda_{i j}^{(t)}\nabla_{\bf{x}_i} g_{i j}\left(\bf{x}_{i}^{(t)}, \bf{x}_{j\to i}^{(t)}\right)\right\|^{2} \notag\\
	&\aleq 2G^2 + 8|\ca{N}_i|\w{G}^2\left\|\@{\lambda}_i^{(t)}\right\|^{2},
\end{align}
where (a) is from Assumption \ref{assump_f} and Proposition \ref{prop}. Part \emph{(b)} of Lemma \ref{Prop_grad_bound} is readily yielded by summing \eqref{Eq_q_i} over $i\in[n]$, i.e., 
\begin{align*} 
	\left\|\bf{q}^{(t)} \right\|^{2} &= \sum_{i=1}^n \left\|\bf{q}_i^{(t)} \right\|^{2} \\
	& \leq \sum_{i=1}^n 2G^2 + 8|\ca{N}_i|\w{G}^2\left\|\@{\lambda}_i^{(t)}\right\|^{2}\\
	& \aleq 2nG^2 + 4m\w{G}^2 \left\|\@{\lambda}^{(t)}\right\|^{2},
\end{align*}
where (a) is because $|\ca{N}_i| \leq \frac{m}{2}$, for all $i\in[n]$. 

The proof of Part \emph{(c)} of Lemma \ref{Prop_grad_bound} is similar to the proof of Part \emph{(a)}. From the definition of $r_{ij}^{(t)}$ in \eqref{Approx_dual_grad}, for any $i\in[n]$ and $j\in\ca{N}_i$, we have
\begin{align*} 
	\left\|\bf{r}^{(t)} \right\|^{2} &= \sum_{i=1}^n \sum_{j\in\ca{N}_{i}} \left(g_{i j}\left(\bf{x}_{i}^{(t)}, \bf{x}_{j\to i}^{(t)}\right) - \delta\eta\lambda_{ij}^{(t)} \right)^2 \\
	& \leq \sum_{i=1}^n \sum_{j\in\ca{N}_{i}} 2 g_{i j}\left(\bf{x}_{i}^{(t)}, \bf{x}_{j\to i}^{(t)}\right)^2 + 2\delta^{2}\eta^{2}\left(\lambda_{ij}^{(t)} \right)^2 \\
	&\aleq 2mC^2 +2 \delta^{2} \eta^{2} \left\|\@{\lambda}^{(t)}\right\|^{2}, 
\end{align*}
where (a) is from Assumption \ref{assump_g}. 

\subsection{Proof of Lemma \ref{Lem_Lag_sum}}

Plugging the results of Lemma \ref{Lem_grad_gap} into \eqref{Eq_Lem_1} and summing over $t\in[T]$, we obtain
\begin{align} \label{Eq_L-L_2}
	&\sum_{t=1}^T \ca{L}^{(t)}\left(\bf{x}^{(t)},\@{\lambda}\right) - \ca{L}^{(t)}(\bf{x},\@\lambda^{(t)})\notag\\
	&\leq \frac{1}{2\eta}\left(\left\|\@{\lambda}-\@{\lambda}^{(1)}\right\|^{2} - \left\|\@{\lambda} - \@{\lambda}^{(T+1)}\right\|^{2}\right) + \frac{\eta}{2}\sum_{t=1}^T \left\|\bf{r}^{(t)}\right\|^{2} \notag\\
	&~~~+ \frac{1}{2\eta}\left(\left\|\bf{x} - \bf{x}^{(1)}\right\|^{2} - \left\|\bf{x} - \bf{x}^{(T+1)}\right\|^{2}\right) + \frac{\eta}{2}\sum_{t=1}^T \left\|\bf{q}^{(t)}\right\|^{2} \notag\\
	&~~~+ \eta T\w{G}^2\left\|\@{\lambda}\right\|^2 + \eta \left(\w{G}^2 + 4R^2L^2 \right) \sum_{t= 1}^{T} \left\|\@{\lambda}^{(t)}\right\|^{2} \notag\\
	&~~~+ \frac{5}{4\eta} \sum_{t= 1}^{T}\sum_{i=1}^n\sum_{j\in\ca{N}_i} \epsilon_{ij}^{(t)}.
\end{align}
In \eqref{Eq_L-L_2}, $\@{\lambda}^{(1)} = \bf{0}$, $\left\|\bf{x} - \bf{x}^{(1)}\right\|^{2} \leq 4R^2$, and the non-positive terms can be omitted. Then, by plugging into the results in Lemma \ref{Prop_grad_bound} and taking expectation over all random variables, Lemma \ref{Lem_Lag_sum} is proved.


\subsection{Proof of Lemma \ref{Prop_error}}
From the definition of $\epsilon_{ij}^{(t)}$ and Eq. \eqref{Eq_prob}, we have
\begin{align*}
	\b{E}\left[\epsilon_{ij}^{(t)}\right] 
	=& p_{ij} \b{E}\left\|\bf{x}_{j}^{(t)} - \bf{x}_{j}^{(t-1)} + \bf{x}_{j}^{(t-1)} -\bf{x}_{j\to i}^{(t-1)} \right\|^2 \notag\\
	\leq& \left(1+\frac{1}{\beta}\right)p_{ij}\b{E}\left\|\bf{x}_{j}^{(t)} - \bf{x}_{j}^{(t-1)}\right\|^2 \notag\\
	& + (1+\beta)p_{ij}\b{E}\left\|\bf{x}_{j}^{(t-1)} -\bf{x}_{j\to i}^{(t-1)}\right\|^2 \notag\\
	\leq& \left(1\!+\!\frac{1}{\beta}\right)p_{ij}\eta^2 \b{E}\left\|\bf{q}_j^{(t-1)} \right\|^2 \!+\! (1\!+\!\beta)p_{ij}\b{E}\left[\epsilon_{ij}^{(t-1)}\right],
\end{align*}
where $\beta>0$ is a constant. Since Algorithm \ref{alg_FI} guarantees $\bf{x}_{j\to i}^{(1)} = \bf{x}_{j}^{(1)}$, we have $\b{E}\left[\epsilon_{ij}^{(1)}\right] = 0$. Then, we have 
\begin{align*} 
	&\b{E}\left[\epsilon_{ij}^{(t)}\right] \leq \left(1+\frac{1}{\beta}\right)p_{ij}\eta^2 \sum_{k=0}^{t-2} ((1+\beta)p_{ij})^k \b{E}\left\|\bf{q}_j^{(t-1-k)} \right\|^2.
\end{align*}
Summing the above equation over $t\in[T]$ yields
\begin{align}  \label{Eq_error2}
	&\sum_{t=1}^T\b{E}\left[\epsilon_{ij}^{(t)}\right] \notag\\
	&\leq \left(1+\frac{1}{\beta}\right)p_{ij}\eta^2 \sum_{t=1}^T \sum_{k=0}^{t-2} ((1+\beta)p_{ij})^k \b{E}\left\|\bf{q}_j^{(t-1-k)} \right\|^2 \notag\\
	& = \left(1+\frac{1}{\beta}\right)p_{ij}\eta^2 \sum_{k= 1}^{T-1} \sum_{t = k+1}^{T} ((1+\beta)p_{ij})^{(t-1-k)} \b{E}\left\|\bf{q}_j^{(k)} \right\|^2 \notag\\
	& \aleq \frac{\left(1+\frac{1}{\beta}\right)p_{ij}\eta^2}{1-(1+\beta)p_{ij}} \sum_{k= 1}^{T-1} \b{E}\left\|\bf{q}_j^{(k)} \right\|^2,
\end{align}
where (a) requires that $(1+\beta)p_{ij}\in[0,1)$, i.e., $\beta\in\left(0, \frac{1}{p_{ij}}-1\right)$. Then, Lemma \ref{Prop_error} is proved by plugging the part \emph{(a)} of Lemma \ref{Prop_grad_bound} into \eqref{Eq_error2}.

\subsection{Proof of Lemma \ref{Lem_band_L-L}}

From the convexity of $\w{\ca{L}}^{(t)}\left(\bf{x},\@\lambda\right)$ w.r.t. $\bf{x}$, we have 
\begin{align*} 
	&\w{\ca{L}}^{(t)}\!\left(\!\bf{x},\@\lambda^{(t)}\!\right)\! \geq \w{\ca{L}}^{(t)}\!\left(\!\bf{x}^{(t)},\@\lambda^{(t)}\!\right)\! +\! \nabla_{\bf{x}}\w{\ca{L}}^{(t)}\!\left(\!\bf{x}^{(t)},\@\lambda^{(t)}\!\right)^\s{T}\!\!\left(\!\bf{x}-\bf{x}^{(t)}\!\right)\!.
\end{align*}
From the concavity of $\w{\ca{L}}^{(t)}\left(\bf{x},\@\lambda\right)$ w.r.t. $\@{\lambda}$, we have 
\begin{align*} 
	&\w{\ca{L}}^{(t)}\!\left(\!\bf{x}^{(t)}, \@{\lambda}\!\right)\! \leq \w{\ca{L}}^{(t)}\!\left(\!\bf{x}^{(t)}, \@{\lambda}^{(t)}\!\right)\! +\! \nabla_{\@{\lambda}} \w{\ca{L}}^{(t)}\!\left(\!\bf{x}^{(t)},  \@{\lambda}^{(t)}\!\right)^\s{T}\! \!\left(\!\@{\lambda} - \@{\lambda}^{(t)}\!\right)\!.
\end{align*}
Combining the above two inequalities yields
\begin{align}
	&\w{\ca{L}}^{(t)}\left(\bf{x}^{(t)},\@{\lambda}\right) - \w{\ca{L}}^{(t)}(\bf{x},\@\lambda^{(t)}) \notag\\
	&\leq  \left\langle \@{\lambda} - \@{\lambda}^{(t)}, \nabla_{\@{\lambda}} \w{\ca{L}}^{(t)}\left(\bf{x}^{(t)},  \@{\lambda}^{(t)}\right) \right\rangle \notag\\
	&~~~+ \left\langle \bf{x}^{(t)} - \bf{x}, \nabla_{\bf{x}} \w{\ca{L}}^{(t)} \left(\bf{x}^{(t)}, \@{\lambda}^{(t)}\right)\right\rangle.  \label{Eq_band_L-L}
\end{align}
Following the same procedures as the proof of Lemma \ref{Lem_L-L} in Appendix C, we have 
\begin{align}\label{Eq_band_lamb}
	&\left\langle \@{\lambda} - \@{\lambda}^{(t)}, \nabla_{\@{\lambda}} \w{\ca{L}}^{(t)}\left(\bf{x}^{(t)},  \@{\lambda}^{(t)}\right) \right\rangle \notag \\
	&\leq \frac{1}{2 \eta}\left(\left\|\@{\lambda} - \@{\lambda}^{(t)}\right\|^{2} - \left\|\@{\lambda} - \@{\lambda}^{(t+1)}\right\|^{2}\right) + \frac{\eta}{2}\left\|\bf{r}^{(t)}\right\|^{2} \notag\\
	&~~~+ \left\langle \@{\lambda} - \@{\lambda}^{(t)}, \nabla_{\@{\lambda}} \w{\ca{L}}^{(t)}\left(\bf{x}^{(t)},  \@{\lambda}^{(t)}\right) - \bf{r}^{(t)}\right\rangle, \\
	&\left\langle \bf{x}^{(t)} - \bf{x}, \nabla_{\bf{x}} \w{\ca{L}}^{(t)}\left(\bf{x}^{(t)}, \@{\lambda}^{(t)}\right)\right\rangle \notag\\
	&\leq \frac{1}{2 \eta}\left(\left\|\bf{x} - \bf{x}^{(t)}\right\|^{2} - \left\|\bf{x} - \bf{x}^{(t+1)}\right\|^{2}\right) + \frac{\eta}{2}\left\|\bf{q}^{(t)}\right\|^{2} \notag\\
	&~~~+ \left\langle \bf{x}^{(t)} - \bf{x}, \nabla_{\bf{x}} \w{\ca{L}}^{(t)}\left(\bf{x}^{(t)}, \@{\lambda}^{(t)}\right) - \bf{q}^{(t)}\right\rangle.  \label{Eq_band_x}
\end{align}
Plugging \eqref{Eq_band_lamb} and \eqref{Eq_band_x} into \eqref{Eq_band_L-L} yields Lemma \ref{Lem_band_L-L}.

\subsection{Proof of Lemma \ref{Lem_band_grad_gap}}

Since $\nabla_{\@\lambda}\w{\ca{L}}^{(t)}\left(\bf{x}^{(t)},\@\lambda^{(t)}\right)= \nabla_{\@\lambda}\ca{L}^{(t)}\left(\bf{x}^{(t)},\@\lambda^{(t)}\right)$, the proof for Part \emph{1)} of Lemma \ref{Lem_band_grad_gap} is the same as the proof for Part \emph{1)} of Lemma \ref{Lem_grad_gap}. The proof for Part \emph{2)} of Lemma \ref{Lem_band_grad_gap} is given below.

Given any $\bf{x}_{i}^{(t)} \in \w{\ca{X}}$, taking expectation over $\w{\bf{q}}_i^{(t)}$ in \eqref{Band_Approx_dual_grad} w.r.t. $ \bf{u}_{i}^{(t)} \sim \ca{U}(\b{S}) $, $\forall i\in[n]$ gives
\begin{align*}
	\b{E}_{\bf{u}_{i}^{(t)}}\left[\left.\w{\bf{q}}_{i}^{(t)}\right|\bf{x}_i^{(t)}\right] \aeq& \nabla_{\bf{x}_{i}} \w{f}_{i}\left(\bf{x}_{i}^{(t)}\right) \\
	&+ 2\sum_{j\in\mathcal{N}_{i}}\lambda_{i j}^{(t)} \nabla_{\bf{x}_{i}} g_{i j}\left(\bf{x}_{i}^{(t)}, \bf{x}_{j\to i}^{(t)}\right), 
\end{align*}
where (a) follows from the expression of $\nabla_{\bf{x}_{i}} \w{f}_{i}\left(\bf{x}_{i}^{(t)}\right)$ given in \eqref{band_primal_grad}. Then, for any $i\in[n]$ and $j\in\ca{N}_i$ we have
\begin{align*}
	& \b{E}_{\bf{u}_{i}^{(t)}}\left[\left.\nabla_{\bf{x}_i} \w{\ca{L}}^{(t)}\left(\bf{x}^{(t)}, \@{\lambda}^{(t)}\right) - \w{\bf{q}}_{i}^{(t)} \right|\bf{x}_i^{(t)}\right] \\
	&= 2\sum_{j\in\mathcal{N}_{i}}\lambda_{i j}^{(t)} \left(\nabla_{\bf{x}_{i}}g_{i j}\left(\bf{x}_{i}^{(t)}, \bf{x}_{j}^{(t)}\right) - \nabla_{\bf{x}_{i}}g_{i j}\left(\bf{x}_{i}^{(t)}, \bf{x}_{j\to i}^{(t)}\right) \right).
\end{align*}
Further, we have 
\begin{align*}
	&\b{E}\left\langle \bf{x}^{(t)} - \bf{x}, \nabla_{\bf{x}} \w{\ca{L}}^{(t)}\left(\bf{x}^{(t)}, \@{\lambda}^{(t)}\right) - \w{\bf{q}}^{(t)}\right\rangle \\
	&= 2\sum_{i=1}^n\sum_{j \in \ca{N}_{i}} \b{E}\left[\lambda_{i j}^{(t)} \left\langle \bf{x}_i^{(t)}-\bf{x}_i, \right.\right.\\
	&~~~\left.\left. \nabla_{\bf{x}_{i}} g_{i j}\left(\bf{x}_{i}^{(t)}, \bf{x}_{j}^{(t)}\right) - \nabla_{\bf{x}_{i}} g_{i j}\left(\bf{x}_{i}^{(t)}, \bf{x}_{j\to i}^{(t)}\right)\right\rangle \right] \\
	&\leq 2\sum_{i=1}^n\sum_{j \in \ca{N}_{i}} \b{E}\left[\lambda_{i j}^{(t)} \left\|\bf{x}_i^{(t)}-\bf{x}_i\right\| \right. \\
	&~~~ \left.\left\| \nabla_{\bf{x}_{i}} g_{i j}\left(\bf{x}_{i}^{(t)}, \bf{x}_{j}^{(t)}\right) - \nabla_{\bf{x}_{i}} g_{i j}\left(\bf{x}_{i}^{(t)}, \bf{x}_{j\to i}^{(t)}\right) \right\| \right] \\
	&\leq 4RL\sum_{i=1}^n\sum_{j \in \ca{N}_{i}} \b{E}\left[\lambda_{i j}^{(t)}\left\|\bf{x}_{j}^{(t)} - \bf{x}_{j\to i}^{t}\right\|\right] \\
	&\leq 4R^2L^2\eta \b{E}\left\|\@{\lambda}^{(t)}\right\|^{2} + \frac{1}{\eta} \sum_{i=1}^n \sum_{j \in \ca{N}_{i}}\b{E}\left\|\bf{x}_{j}^{(t)} -\bf{x}_{j\to i}^{(t)} \right\|^2, 
\end{align*}
which proves Lemma \ref{Lem_band_grad_gap}.

\subsection{Proof of Lemma \ref{Prop_band_grad_bound}}

We first prove Part \emph{(a)} of Lemma \ref{Prop_band_grad_bound}. From \eqref{Band_Approx_dual_grad}, we have 
\begin{align} \label{Eq_band_q_i}
	\left\|\w{\bf{q}}_i^{(t)} \right\|^{2} &= \left\|\frac{d}{2 \zeta}\left[f_{i}\left(\bf{x}_{i}^{(t)}\!+\!\zeta \bf{u}_{i}^{(t)}\right)-f_{i}\left(\bf{x}_{i}^{(t)}\!-\!\zeta \bf{u}_{i}^{(t)}\right)\right] \bf{u}_{i}^{(t)} \right. \notag\\
	&~~~\left.+ \sum_{j\in\ca{N}_{i}} 2\lambda_{i j}^{(t)} \nabla_{\bf{x}_{i}} g_{i j}\left(\bf{x}_{i}^{(t)}, \bf{x}_{j\to i}^{(t)}\right)\right\|^{2} \notag\\
	\aleq& 2\left(\frac{d}{2\zeta} 2\zeta G \left\|\bf{u}_{i}^{(t)}\right\|^2 \right)^2 + 8|\ca{N}_i| \sum_{j\in\ca{N}_{i}}\left(\lambda_{i j}^{(t)}\right)^{2}\w{G}^2\notag\\
	\bleq & 2 d^2G^2 + 8|\ca{N}_i|\w{G}^2\left\|\@{\lambda}_i^{(t)}\right\|^{2},
\end{align}
where (a) is from Assumption \ref{assump_f} and Proposition \ref{prop}, and (b) is because $\bf{u}_{i}^{(t)} \sim \mathcal{U}(\mathbb{S})$. 
Then, Part \emph{(b)} of Lemma \ref{Prop_band_grad_bound} is readily obtained by summing \eqref{Eq_band_q_i} over $i\in[n]$, i.e.,
\begin{align*} 
	\left\|\w{\bf{q}}^{(t)} \right\|^{2} &= \sum_{i=1}^n \left\|\w{\bf{q}}_i^{(t)} \right\|^{2} \\
	& \leq \sum_{i=1}^n  2 d^2G^2 + 8|\ca{N}_i|\w{G}^2\left\|\@{\lambda}_i^{(t)}\right\|^{2} \\
	& \aleq 2nd^2G^2 + 4m\w{G}^2 \left\|\@{\lambda}^{(t)}\right\|^{2},
\end{align*} 
where (a) is because $|\ca{N}_i| \leq \frac{m}{2}$ for all $i\in[n]$. The proof of Part \emph{(c)} in Lemma \ref{Prop_band_grad_bound} is the same as the proof of Part \emph{(c)} in Lemma \ref{Prop_grad_bound}, given in Appendix E. 

\subsection{Proof of Lemma \ref{Lem_band_Lag_sum}}

Plugging the results in Lemma \ref{Lem_band_grad_gap} into Lemma \ref{Lem_band_L-L} and summing over $t\in[T]$, we obtain
\begin{align} \label{Eq_band_L-L_2}
	&\sum_{t=1}^T \b{E}\left[\w{\ca{L}}^{(t)}\left(\bf{x}^{(t)},\@{\lambda}\right) - \w{\ca{L}}^{(t)}(\bf{x},\@\lambda^{(t)}) \right] \notag\\
	&\leq \frac{1}{2\eta}\left(\left\|\@{\lambda}-\@{\lambda}^{(1)}\right\|^{2} - \left\|\@{\lambda} - \@{\lambda}^{(T+1)}\right\|^{2}\right) + \frac{\eta}{2}\sum_{t=1}^T \b{E}\left\|\bf{r}^{(t)}\right\|^{2} \notag\\
	&~~~+ \frac{1}{2\eta}\left(\left\|\bf{x}\!-\!\bf{x}^{(1)}\right\|^{2} - \left\|\bf{x} - \bf{x}^{(T+1)}\right\|^{2}\right) + \frac{\eta}{2}\sum_{t=1}^T \b{E}\left\|\bf{q}^{(t)}\right\|^{2} \notag\\
	&~~~+ \eta T\w{G}^2\left\|\@{\lambda}\right\|^2 + \eta \left(\w{G}^2 + 4R^2L^2 \right) \sum_{t= 1}^{T} \b{E}\left\|\@{\lambda}^{(t)}\right\|^{2} \notag\\
	&~~~+ \frac{5}{4\eta} \sum_{t= 1}^{T}\sum_{i=1}^n\sum_{j\in\ca{N}_i}\b{E}\left[\epsilon_{ij}^{(t)}\right].
\end{align}
In \eqref{Eq_band_L-L_2}, $\@{\lambda}^{(1)} = \bf{0}$, $\left\|\bf{x} - \bf{x}^{(1)}\right\|^{2} \leq 4R^2$, and the non-positive terms can be omitted. We take $\bf{x}$ in \eqref{Eq_band_L-L_2} as $(1-\alpha)\bf{x}^* + \alpha \o{\bf{z}}_0$, where $\o{\bf{z}}_0:=[\bf{z}_0^\s{T},\cdots,\bf{z}_0^\s{T}]^\s{T}\in\b{R}^{nd}$. Then, by plugging into the results in Lemma \ref{Prop_band_grad_bound}, Lemma \ref{Lem_band_Lag_sum} is proved.


\subsection{Proof of Lemma \ref{Prop_band_error}}

Referring to the procedures in Appendix G, we have 
\begin{align}  \label{Eq_band_error2}
	\sum_{t=1}^T\b{E}\left[\epsilon_{ij}^{(t)}\right]&\leq \frac{\left(1+\frac{1}{\beta}\right)p_{ij}}{1-(1+\beta)p_{ij}}\eta^2 \sum_{k= 1}^{T-1} \b{E}\left\|\w{\bf{q}}_j^{(k)} \right\|^2,
\end{align}
which can be obtained by replacing $\bf{q}_i^{(t)}$ in Appendix G with $\w{\bf{q}}_i^{(t)}$, $\forall (i, j)\in\ca{E}$.
Then, plugging Part \emph{(a)} of Lemma \ref{Prop_band_grad_bound} into \eqref{Eq_band_error2}, we have
\begin{align*}
	\sum_{t=1}^T \b{E}\left[\epsilon_{ij}^{(t)}\right] 
	\leq& \frac{2\left(1+\frac{1}{\beta}\right)p_{ij}}{1-(1+\beta)p_{ij}}\eta^2 Td^2G^2  \\
	&+ 8\eta^2\w{G}^2 \frac{\left(1+\frac{1}{\beta}\right)p_{ij}}{1-(1+\beta)p_{ij}} \sum_{t= 1}^{T} \b{E}\left\|\@{\lambda}_j^{(t)}\right\|^{2}.
\end{align*}
Summing the above inequality over $i\in[n]$ and $j\in\ca{N}_i$ proves Lemma \ref{Prop_band_error}.

\bibliographystyle{IEEEtran}
\bibliography{pairwise}



\end{document}